\pdfoutput=1

\documentclass[11pt]{article}

\usepackage[]{EACL2023}

\usepackage{times}
\usepackage{latexsym}

\usepackage[T1]{fontenc}

\usepackage[utf8]{inputenc}

\usepackage{microtype}

\usepackage{inconsolata}

\usepackage{amssymb, bm}
\usepackage{amsmath}
\usepackage{multirow}
\usepackage{booktabs} 
\usepackage{graphicx}
\usepackage{soulutf8}
\usepackage{subfig}
\providecommand{\methodName}{\mbox{\it Value Zeroing}}

\title{Quantifying Context Mixing in Transformers}

\author{
    Hosein Mohebbi$^{1}$ ~ Willem Zuidema$^{2}$ ~ Grzegorz Chrupała$^{1}$  ~ Afra Alishahi$^{1}$ \\
    $^1$ CSAI, Tilburg University ~
    $^2$ ILLC, University of Amsterdam \\
    \texttt{\{h.mohebbi, a.alishahi\}@tilburguniversity.edu}\\
    \texttt{w.h.zuidema@uva.nl}\\
    \texttt{grzegorz@chrupala.me}\\
  }

\begin{document}
\maketitle

\begin{abstract}
Self-attention weights and their transformed variants have been the main source of information for analyzing token-to-token interactions in Transformer-based models. But despite their ease of interpretation, these weights are not faithful to the models' decisions as they are only one part of an encoder, and other components in the encoder layer can have considerable impact on information mixing in the output representations. In this work, by expanding the scope of analysis to the whole encoder block, we propose \mbox{\emph{\methodName}}, a novel context mixing score customized for Transformers that provides us with a deeper understanding of how information is mixed at each encoder layer. We demonstrate the superiority of our context mixing score over other analysis methods through a series of complementary evaluations with different viewpoints based on linguistically informed rationales, probing, and faithfulness analysis.\footnote{Code is freely available at \url{https://github.com/hmohebbi/ValueZeroing}} 
\end{abstract}

\section{Introduction}
Transformers \citep{NIPS2017_3f5ee243}, with their impressive empirical success, have become a prime choice of architecture to learn contextualized representations across a wide range of modalities, such as language \citep{devlin-etal-2019-bert, NEURIPS2020_1457c0d6}, vision \citep{dosovitskiy2021an}, vision-language \cite{Radford2021LearningTV, Rombach2022HighResolutionIS}, and speech \citep{baevski2020wav2vec}, mainly due to their ability to utilize pairwise interactions between input tokens at every timestep.

To better understand the inner dynamics of Transformers, we need to trace the information flow from the input embeddings up to the output representation (including quantifying the degree of {\it context mixing}, which we will define below).
The attention weights from the multi-head attention mechanisms offer a straightforward starting point for understanding this flow, 
and these weights (`raw attention') have been used in many studies \citep[\textit{inter alia}]{clark-etal-2019-bert, kovaleva-etal-2019-revealing, reif2019visualizing, bert-track-syntactic-dep}. However, the reliability and usefulness of raw attention weights has also been questioned \citep{jain-wallace-2019-attention, bibal-etal-2022-attention}.
In particular, attention weights tend to concentrate on uninformative tokens in the context \citep{voita-etal-2018-context, clark-etal-2019-bert}, and
removing or altering them may lead to the same and sometimes even better model performance on downstream tasks \citep{jain-wallace-2019-attention, NEURIPS2019_749a8e6c, hassid-etal-2022-much}. These findings suggest that Transformers do not solely rely on self-attention, and other components in the encoder block play an essential role in information mixing.

A number of methods have been proposed to compute some form of `effective attention weights', with the goal of more faithfully tracing the relative contributions of different input tokens at various layers of the Transformer (as we will discuss in Section~\ref{sec:relatedwork}).
These methods show an improvement over raw attention, but still ignore key components of the Transformer encoder block. This is a particularly crucial shortcoming, given that most of the parameter budget in a Transformer encoder is spent on position-wise feed-forward networks outside of the self-attention component, which can have a considerable impact on the degree of information mixing in the output representations.

In this paper, we focus on \emph{context mixing}: the property of Transformers that in each node, at each layer, information from the context can be incorporated into the representation of the target token. 
We propose {\methodName}, a novel approach to quantify the contribution each context token has in determining the final representation of a target token, at each layer of a Transformer. 
{\methodName} is based on the Explaining-by-Removing intuition \citep{Covert2021ExplainingBR} shared by many posthoc interpretability methods, but it takes advantage of a specific feature of Transformers: it zeroes only the value vector of a token $t$ when computing its importance, but leaves the key and query vectors (and thus the pattern of information flow) intact.

Based on extensive experiments and three complementary approaches to evaluation, we demonstrate that importance scores we can obtain with {\methodName} provide better interpretations than other analysis methods. 

Firstly, we use a set of grammatical agreement tasks from the BLiMP corpus \citep{warstadt-etal-2020-blimp} as a case study. Transformer-based models do extremely well on the task of distinguishing grammatical from ungrammatical sentences, and the BLiMP corpus provides information on the \emph{cue} words that determine the difference. We find that {\methodName}, unlike earlier approaches, indeed reveals that Transformers make use of relevant cues. 

Secondly, we use information-theoretic probing \citep{voita-titov-2020-information,pimentel-etal-2020-information} as an independent approach to track information flow in Transformer networks.
The scores we obtain with {\methodName} turn out to be highly correlated with layer-wise probing performance; that is, probing accuracy is higher in layers where relevant tokens are more effectively utilized by the model. 

Thirdly, we assess the faithfulness of our method \citep{jacovi-goldberg-2020-towards}; compared to alternative analysis methods, we show that {\methodName} is not only more plausible and human-interpretable, but also more faithful to models' decisions.

\section{Related Work}
\label{sec:relatedwork}

While numerous studies have leveraged the weights assigned by the self-attention mechanism to gain intuition about the information mixing process in Transformers \citep{clark-etal-2019-bert, kovaleva-etal-2019-revealing, reif2019visualizing, lin-etal-2019-open, Htut2019DoAH, raganato-tiedemann-2018-analysis}, it is still a matter of debate whether attention weights are suitable for interpreting the model (see \citet{bibal-etal-2022-attention}'s study for a full discussion). 
Thus several post-processing interpretability techniques have been proposed to convert these weights into scores that provide a more
detailed interpretation of the inner workings of Transformers. We review the main approaches below.

\citet{abnar-zuidema-2020-quantifying} propose the \emph{attention-flow} and \emph{attention-rollout} methods to approximate information flow in Transformers based on raw attention weights. The former treats raw attention weight matrices as a flow network and returns the maximum flow through each input token. The latter recursively multiplies the attention weight matrix at each layer by the preceding ones. There is, however, an unjustified assumption in the formulation of these methods that both multi-head attention and residual connections contribute equally to the computation of the output. 

\citet{kobayashi-etal-2020-attention} propose a method that incorporates the norm of the transformed value vectors and report a negative correlation between these norms and raw attention weights on frequent tokens, which partially explains the insufficiency of raw attention weights for context mixing estimation. \citet{kobayashi-etal-2021-incorporating} extend this method to the whole self-attention block by incorporating Residual connections (RES) and Layer Normalization (LN) (two components with significant impact on both model performance and training convergence \cite{pmlr-v119-parisotto20a, liu-etal-2020-rethinking}), but demonstrate that RES and LN components largely cancel out the mixing process.
\citet{kobayashi-etal-2021-incorporating}'s method, however, ignores the effect of the second sublayer in a Transformer's encoder.

\citet{Brunner2020On} and \citet{pascual-etal-2021-telling} employ a gradient-based approach for analyzing the interaction of input representations, but the gradient measures the sensitivity between two vectors and ignores the impact of the input vector. 
In our experiments we show that despite their relative success in explaining model decisions, gradient-based approaches are not suitable for layer-wise analysis.

More recently, the effectiveness of combining these approaches has also been investigated.  
\citet{modarressi-etal-2022-globenc} propose a method that uses \citet{kobayashi-etal-2021-incorporating}'s scores and  incorporates the effects of the second layer normalization; they aggregate those scores using \emph{rollout} \citep{abnar-zuidema-2020-quantifying} to provide global token attributions.
In the same vein, \citet{ferrando-etal-2022-measuring} use rollout to aggregate a variant of \citet{kobayashi-etal-2021-incorporating}'s scores: instead of relying on the Euclidean norm of the transformed vector, they measure Manhattan distance of each transformed vector to the context vector outputted from self-attention block. In both studies, however, the fact that these context vectors might undergo significant changes after passing through the second sublayer in the encoder layer is not taken into account.
We will show (in Section~\ref{sec:faithfulness}) that even at a global level,
our scores provide better interpretation than prior methods.

\section{Our Proposed Method}
To remedy for the limited scope of the existing methods, we introduce a new context mixing score that takes into account all components in a Transformer encoder block. 
\subsection{Background and Notation}
\label{sec:background}
In this section, we set up the notation and briefly review the internal structure of an encoder layer in the Transformer architecture.

Each Transformer encoder layer is composed of two sublayers: a multi-head self-attention mechanism (\textbf{MHA}) and a position-wise fully connected feed-forward network (\textbf{FFN}), followed by a Residual connection (\textbf{RES}) and Layer Normalization (\textbf{LN}) around each of these two sublayers. This encoder layer produces the next contextualized representations $(\bm{\tilde{x}}_1, ..., \bm{\tilde{x}}_n)$ for each token in the context, using the output representations from the previous layer $(\bm{x}_1, ..., \bm{x}_n)$.

\paragraph{MHA.}
For each head $h \in \{1,...,H\}$ in the self-attention module, each input vector $\bm{x}_i$ is transformed into a query $\bm{q}_i^h$, a key $\bm{k}_i^h$, and a value $\bm{v}_i^h$ vector via separate trainable linear transformations:
\vspace{-0.3cm}
\begin{equation}
\label{eq:query}
{\bm{q}_i^h = \bm{x}_i \bm{W_Q^h} + \bm{b}_Q^h}
\end{equation}
\vspace{-0.3cm}
\begin{equation}
\label{eq:key}
{\bm{k}_i^h = \bm{x}_i \bm{W_K^h} + \bm{b}_K^h}
\end{equation}
\vspace{-0.3cm}
\begin{equation}
\label{eq:value}
{\bm{v}_i^h = \bm{x}_i \bm{W_V^h} + \bm{b}_V^h}
\end{equation}
The context vector $\bm{z}_i^h$ for the $i^\text{th}$ token of each attention head is then generated as a weighted sum over the transformed value vectors:
\begin{equation}
\label{eq:context_vectors}
\bm{z}_i^h = \sum_{j=1}^{n}\alpha_{i,j}^h \bm{v}^h_j
\end{equation}
where $\alpha_{i,j}^h$ is the raw attention weight assigned to the $j^\text{th}$ token, and computed as a softmax-normalized dot product between the corresponding query and key vectors:
\begin{equation}
\label{eq:raw_attentions}
\alpha_{i,j} = \mathop{\rm{softmax}}_{\bm{x}_j \in \mathcal{X}}\left(\frac{\bm{q}_i \bm{k}_j^\top}{\sqrt{d}} \right) \in \mathbb{R}
\end{equation}
Next, the context vector ($\bm{z}_i \in \mathbb{R}^d$) for the $i^\text{th}$ token is computed by concatenating all the heads' outputs followed by a head-mixing $\bm{W}_O$ projection and layer normalization:
\begin{equation}
\label{eq:concatenated_context_vectors}
\bm{z}_i = \textsc{Concat}(\bm{z}_i^1, ..., \bm{z}_i^H)\bm{W}_O
\end{equation}
\vspace{-0.3cm}
\begin{equation}
\label{eq:normalized_concatenated_context_vectors}
    \bm{z}_i = {\rm LN_\text{MHA}}(\bm{z}_i + \bm{x}_i)
\end{equation}
\paragraph{FFN.}
Each encoder layer also includes two linear transformations with a ReLU activation in between, which is applied to every $\bm{z}_i$ separately and identically to produce output token representations $\bm{\tilde{x}}_i$:
\begin{equation}
\label{eq:FFN}
\bm{\tilde{x}}_i = {\rm max}(0, \bm{z}_i \bm{W}_1 + \bm{b}_1) \bm{W}_2 + \bm{b}_2 
\end{equation}
\vspace{-0.3cm}
\begin{equation}
\label{eq:normalized_FFN}
    \bm{\tilde{x}}_i = {\rm LN_\text{FFN}}(\bm{\tilde{x}}_i + \bm{z}_i)
\end{equation}

\begin{table*}[t]
\centering
\resizebox{0.95\linewidth}{!}{
\begin{tabular}{lllll}
\toprule
{\bf Phenomenon}  & {\bf UID} & {\bf Example} & {\bf Target word} & {\bf Foil word} \\ 
\midrule
\multirow{1}{*}{Anaphor Number Agreement} & ana & \ul{Many teenagers} were helping [MASK]. & themselves & herself \\
\midrule
\multirow{2}{*}{Determiner-Noun Agreement} & dna & Jeffrey has not passed [MASK] \ul{museums}. & these & this\\
 & dnaa & Sara noticed [MASK] white \ul{hospitals}. & these & this \\
\midrule
\multirow{2}{*}{Subject-Verb Agreement} & darn & The \ul{pictures} of Martha [MASK] not disgust Anne. & do & does \\
& rpsv & \ul{Kristen} [MASK] fixed this chair. & has & have\\
\bottomrule
\end{tabular}
}
\caption{Examples of the selected tasks with our annotations from the BLiMP benchmark (UIDs are unique identifiers used in BLiMP). Cue words are \ul{underlined}.} 
\label{Tab:blimp}
\end{table*}

\subsection{Value Zeroing}
We aim to measure how much a token uses other context tokens to build its output representation $\bm{\tilde{x}}_i$ at each encoder layer.
To this end, we treat the self-attention mechanism as a fuzzy hash-table, where we look up the sum of values weighted by the query-key match in the context.
Thus in Eq.~\ref{eq:context_vectors} we replace a value vector associated with token $j$ with a zero vector \mbox{$\bm{v}_j^h \gets \bm{0}, \forall h \in H$}, where the context vector for the $i^\text{th}$ token is being computed. This provides an alternative output representation $\bm{\tilde{x}}_i^{\neg j}$ for the $i^\text{th}$ token that has excluded token $j$ in the mixing process.
By comparing the alternative output representation $\bm{\tilde{x}}_i^{\neg j}$ with the original $\bm{\tilde{x}}_i$, we can measure how much the output representation is affected by the exclusion of the $j^\text{th}$ token:
\begin{equation}
\label{eq:dependency_score}
    \mathcal{C}_{i,j} = \bm{\tilde{x}}_i^{\neg j} \ast \bm{\tilde{x}}_i
\end{equation}
where the operation $\ast$ can be any pairwise distance metric that properly considers the characteristics of the model's representation space. 
We opted for \mbox{\emph{cosine}} distance throughout our experiments as its superiority over other dissimilarity metrics has been supported for textual deep learning models \citep{yokoi-etal-2020-word, hanawa2021evaluation}.\footnote{More details on the choice of distance metric is discussed in Appendix \ref{appendix:more_distance_functions}.}
Computing Eq.~\ref{eq:dependency_score} for all $\bm{\tilde{x}}_i$ in a given context provides us with a \emph{{\methodName}} matrix score $\bm{\mathcal{C}}$ where the value of the cell $\mathcal{C}_{i,j} \in \mathbb{R}$ ($i^\text{th}$ row, $j^\text{th}$ column) in the map denotes the degree to which the $i^\text{th}$ token depends on the $j^\text{th}$ token to form its contextualized representation.

Note that unlike generic perturbation approaches, our proposed method \emph{does not} remove the token representations $\bm{x}_i$ from the input of an encoder. We argue that ablating input token representations cannot be a reliable basis to understand context mixing process since any changes in the input vectors will lead to changes in the query and key vectors (Eq.~\ref{eq:query} and \ref{eq:key}), resulting in a change in the attention distribution (cf.\ Eq.~\ref{eq:raw_attentions}). Consequently, there will be a discrepancy between the alternative attention weights and those for the original context. 
Instead, our method only nullifies the \emph{value vector} of a specific token representation. In this way, the token representation can maintain its identity within the encoder layer, but it does not contribute to forming other token representations.
Moreover, since our {\methodName} is computed from the encoder's layer outputs, it incorporates all the components inside an encoder layer such as multi-head attention, residual connection, layer normalization, and also feed-forward networks, resulting in a more reliable context mixing score than previous methods.

\section{Experimental Setup}
\subsection{Data}
We used the BLiMP benchmark \citep{warstadt-etal-2020-blimp} which contains a set of pairs of minimally different sentences that contrast in grammatical acceptability under a specific linguistic phenomenon. The benchmark isolates linguistic phenomena such that only one word determines the true label of each sentence. We refer to this crucial context token as the {\it cue word}. The nature of this task makes it especially suitable for evaluating context mixing scores, since it gives us a strong hypothesis on which context token is the most relevant for the representation of the masked {\it target word}.

From this benchmark, we select five datasets with three different linguistic phenomena for which Pre-trained Language Models (PLMs) have shown high accuracy to ensure that the model captures the relevant information. We expand contractions such as \emph{doesn't} $\to $ \emph{does not}) and generate dependency trees using SpaCy \citep{spacy2} to extract and annotate the position of target and cue words in a sentence. In Table~\ref{Tab:blimp}, we provide an example of each phenomenon in the benchmark together with our automated annotations. 
We accumulate examples from the five selected tasks as a unified dataset for grammatical agreement, resulting in 4,276 data points, and divide them equally into Train and Test sets. The Train set is only used for the fine-tuning phase; the Test set is used for all evaluation experiments.

\subsection{Target Model}
We conduct our experiments on three Transformer-based language models: BERT \citep[uncased,][]{devlin-etal-2019-bert}, RoBERTa \citep{Liu2019RoBERTaAR} and ELECTRA \citep{Clark2020ELECTRA:}.\footnote{Base, with 12 layers and 12 attention heads, obtained from the Transformers library \citep{wolf-etal-2020-transformers}.} The results for the latter two are reported in Appendix~\ref{appendix:more_plms}. By replacing the target words with the [MASK] token, we perform a Masked Language Modeling (MLM) task using the model's pre-trained MLM head. For instance, in the Subject-Verb Agreement example \emph{\text{``The pictures of Martha \ul{do} not disgust Anne.''}}, we replace the verb \emph{`do'} with the [MASK] token and feed the example to the model. 

We perform our experiments on both pre-trained and fine-tuned versions of each model. Including a fine-tuned model in our analysis study gives us a complementary insight into the importance of the cue words, since fine-tuning allows the model to concentrate on the most helpful words for the downstream task of choice (i.e., agreement) and makes sure that target word representations take the cue word into account. We use prompt fine-tuning \citep{schick-schutze-2021-exploiting, schick-schutze-2021-just, karimi-mahabadi-etal-2022-prompt} and compute Cross Entropy loss only over the output logits corresponding to the target and foil classes. Accuracy is $0.96$ for pre-trained and $0.99$ for fine-tuned BERT.

\subsection{Baselines}

Here we describe the existing context mixing methods which we include in our experiments. For each method, we select the $m^\text{th}$ row of its context mixing map where $m$ is the position of the [MASK] token, resulting in a 1-D array of scores for each context token. We normalize the scores for all tokens in a sentence so that they are all positive values and sum to one. For completeness, we also include a few  gradient-based attribution methods in our comparisons.

\vspace{.2cm}\noindent{\bf Rand:} random scores generated from a uniform distribution for each sentence in the dataset.

\vspace{.2cm}\noindent{\bf Attn:} raw attention scores $\bm{\alpha}_{m}$ from Eq.~\ref{eq:raw_attentions}.

\vspace{.2cm}\noindent{\bf Attn-rollout:} An aggregation method for approximating the attention flow based on raw attention weights \citep{abnar-zuidema-2020-quantifying}.

\vspace{.2cm}\noindent{\bf Attn-Norm:} norm-based method of \citet{kobayashi-etal-2020-attention} that also incorporates the norm of the transformed input vectors to compute context mixing scores.

\vspace{.2cm}\noindent{\bf Attn-Norm + RES + LN:} the extended norm-based method of \citet{kobayashi-etal-2021-incorporating} in which they also incorporates Residual connection (RES) and Layer Normalization (LN) located only in the first sublayer of a Transformer's encoder.

\vspace{.2cm}\noindent{\bf GlobEnc \& ALTI:}
Two global token attribution methods proposed by \citet{modarressi-etal-2022-globenc} and \citet{ferrando-etal-2022-measuring}. 
For a fair comparison, we exclude the aggregation through rollout in our first two evaluations (referred to as GlobEnc$\neg$ and ALTI$\neg$), because we are interested in assessing context mixing scores at each layer separately. In our third evaluation, we compare these methods with others at a global level and use rollout.

\vspace{.2cm}\noindent{\bf GradXinput, IG \& DL:} feature attribution scores that also make use of \emph{top-down} information from the classification layer on top of the Transformer. We consider three popular variants of gradient-based attribution scores: Gradient$\times$Input (\textbf{GradXinput}) \citep{samek2019explainable, 10.1609/aaai.v33i01.33015717}, Integrated Gradients (\textbf{IG}) \citep{10.5555/3305890.3306024}, and DeepLift (\textbf{DL}) \citep{10.5555/3305890.3306006}.
For gradient-based methods, we use the pre-trained MLM head which has been trained during the pre-training of the BERT to compute the gradient of the true label with respect to the token representations at each layer.

\begin{figure*}[h!]
\centering
    \includegraphics[width=0.94\linewidth]{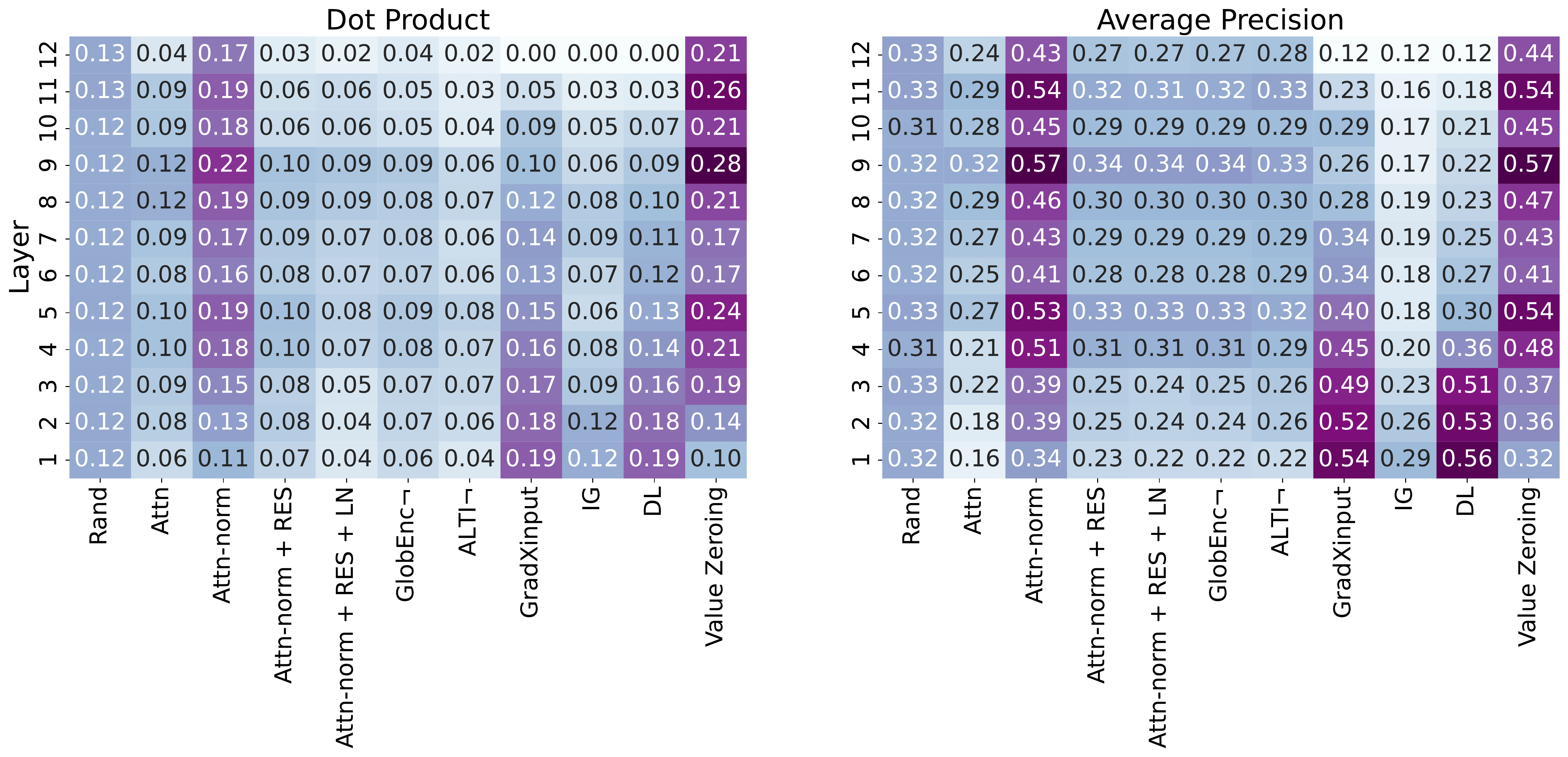}
    \caption{Layer-wise alignment between the cue vector and different analysis methods averaged over Test set examples for the \textbf{pre-trained} model. Higher value (darker color) is better.} 
    \label{fig:layerwise_rationales_pt}
\end{figure*}
\begin{figure*}[h!]
\centering
    \includegraphics[width=0.88\linewidth]{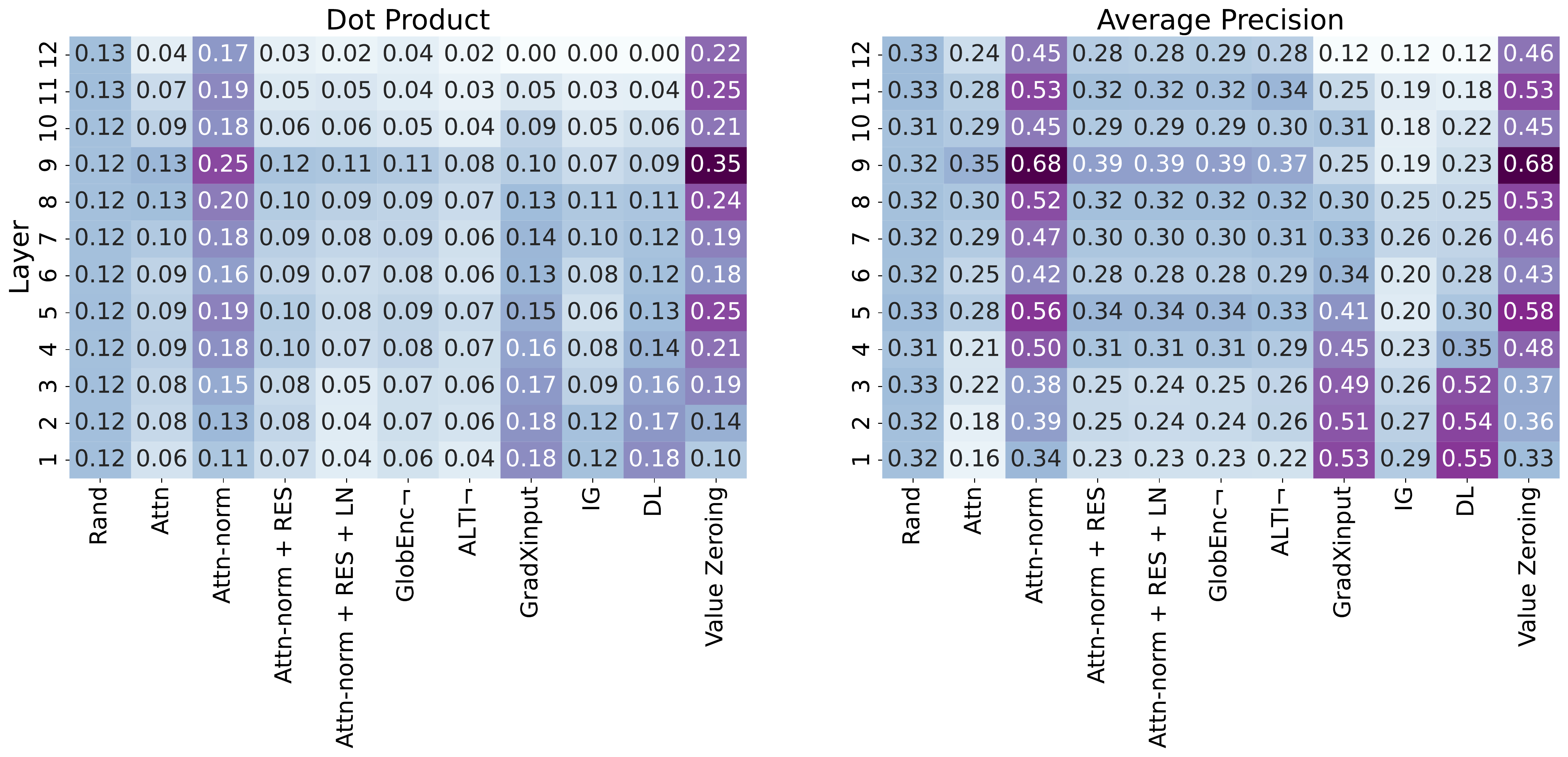}
    \caption{Layer-wise alignment between the cue vector and different analysis methods averaged over Test set examples for the \textbf{fine-tuned} model. Higher value (darker color) is better.}
    \label{fig:layerwise_rationales_ft}
\end{figure*}

\section{Evaluation 1: Cue Alignment}
\label{sec:rationale_layerwise_results}
As cue words are the only indicators of the true labels in our dataset, we expect that when the model performs well, it overwhelmingly depends on these words to form the representation of a [MASK] token in a given context. 
To quantify the alignment between a context mixing score and the cue word, we first define a binary cue vector $\xi$ according to the following condition:
\begin{equation}
\xi_i = 
  \begin{cases}
    1, & \textrm{the $i^\text{th}$ token} \in \textrm{Cue words} \\
    0, & \textrm{otherwise} \
  \end{cases}
  \label{eq:rationale_vector}
\end{equation}
Then we compare the cue vector and the prediction of a context mixing score $\bm{S}$ in two different ways:
\paragraph{Dot Product.}
We quantify cue alignment as $\bm{S}\cdot\bm{\xi}$, which measures the total score mass the model assigns to cue words to form the representation of the target token.
\paragraph{Average Precision.} We quantify cue alignment as the average precision between the two vectors,
which is a weighted mean of precision at each recall level:
\begin{equation}
  \label{eq:ap}
  \text{AP} = \sum_n (R_n - R_{n-1}) P_n
\end{equation}
where $P_n$ and $R_n$ are the precision and recall at the $n^\text{th}$ threshold. This metric relies on the ranking of tokens rather than the magnitude of their weights.\footnote{We also employed Probes-needed \citep{Zhong2019FinegrainedSA} metric in our evaluation which intuitively counts the number of non-cue tokens we need to probe to find cue words based on a given score. As its motivation is similar to Average Precision and the results show the same pattern, we relegate the results with this metric to the Appendix \ref{appendix:more_metrics}.}

Figures \ref{fig:layerwise_rationales_pt} and \ref{fig:layerwise_rationales_ft} show the alignment between the cue vector and different analysis methods using dot product and average precision for the pre-trained and fine-tuned model, respectively. In comparison with the other context mixing methods, {\methodName} shows a higher degree of the target model incorporating cue words into the representation of the [MASK] token across all layers.

As can be seen from the first two columns in all graphs, raw self-attention weights (\textbf{Attn}) always perform worse than even random scores in highlighting cue words. This is in line with previous studies showing that raw attention weights often pay attention to uninformative tokens \citep{voita-etal-2018-context, clark-etal-2019-bert} and do not reflect the appropriate context  \mbox{\citep{kim-etal-2019-document}}.
%
However, we can see a significant improvement in the results for \textbf{Attn-norm} where the norm of transformed value vectors are also taken into account, confirming that value vectors play an essential role in the context mixing process.
The method \textbf{Attn-norm + RES + LN}, which expands \textbf{Attn-norm} to the whole self-attention block by adding Residual connection and Layer Normalization, would seem to show that the model is incapable of utilizing the cue words. However, incorporating also the second part of the encoder layer via our method shows the model does indeed use the cue words.
We also find that GlobEnc and ALTI
benefit from the rollout aggregation method to provide a global view, but they do not seem to provide good layerwise scores (without rollout).

The gradient-based scores, in contrast to the other methods, highlight the cue words only in the earlier layers of the model. In the next section, by using a layer-wise probing experiment, we will show that these scores are not reliable for identifying the relevant context in individual layers.

\section{Evaluation 2: Context Mixing versus Probing}
\label{sec:probing}
In this section, we investigate the relationship between cue word alignment and probing performance across layers. 
We hypothesize that if a layer aligns better with the cue word according to a reliable context mixing score, then the representation of the masked token on that layer can be used more effectively by a probing classifier to decode number agreement with the cue word. 

To verify our hypothesis, we obtain the representation of masked tokens in test examples across all layers. Since all examples in our dataset share the same number agreement property, we associate each masked representation with a \emph{Singular} or \emph{Plural} label.
Next, we perform an information-theoretic probing analysis using Minimum Description Length (MDL) to measures the degree to which representations encode number agreement.
We chose MDL as our probe since it is theoretically justified and has been shown to provide more reliable results than conventional probes \citep{voita-titov-2020-information, fayyaz-etal-2021-models}.

To compute MDL, we employed the \emph{online coding} of \citet{voita-titov-2020-information}.
Since MDL can be affected by the number of data points ($\text{N}$), we 
measure compression as our evaluation metric which is defined as follows:
\begin{equation}
\text{Compression} = \frac{N\cdot\log_2(K)}{\text{MDL}}
\label{eq:compression}
\end{equation}
where $K$ refers to the number of classes (2 in our case).
This metric is equal to 1 (no compression) for a random guessing classifier. A higher value for Compression indicates more accurate label prediction for the probing classifier.

\begin{figure}[t]
    \centering
    \includegraphics[scale=0.55]{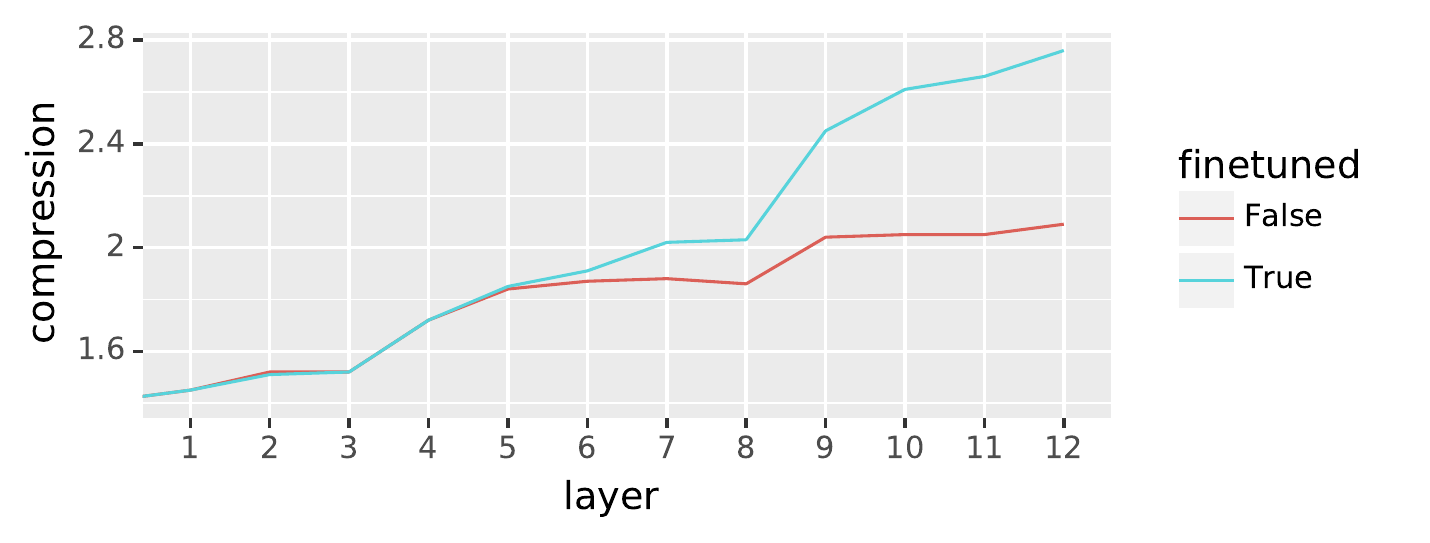}
    \caption{Layer-wise compression of probing classifiers using pre-trained and fine-tuned representations.}
\label{fig:probing}
\end{figure}

Figure~\ref{fig:probing} reports compression of probing classifiers based on representations obtained from both pre-trained and fine-tuned models across all layers. 
We also include the results for the embedding layer of the model (layer 0) which can serve as a non-contextualized baseline.
We can see a jump in probing performance at layers 4 ($1.52 \to 1.72$) and 9 ($2.03 \to 2.45$) in the fine-tuned setup, the same layers for which we found a higher alignment with cue words in Figures~\ref{fig:layerwise_rationales_pt} and \ref{fig:layerwise_rationales_ft}.

Table \ref{tab:corr_probing_rationale} presents the correlation between layer-wise Compression scores and layer-wise cue alignment scores from Section~\ref{sec:rationale_layerwise_results} for different analysis methods.
As we can see, alignment according to {\methodName} is highly positively correlated with the probing performance. This suggests that when {\methodName} indicates that the model uses cue words to form representations of the masked tokens in a particular layer, these representations are in fact better at encoding number agreement.

Recall that based on Figures~\ref{fig:layerwise_rationales_pt} and \ref{fig:layerwise_rationales_ft}, according to the gradient-based methods the masked tokens pay more attention to the cue words only in earlier layers. However, we can see a highly negative correlation with probing results for these scores. Due to the nature of the task the probing score goes up monotonically along the layers. At the same time, the gradient attribution score goes up monotonically as you get closer to the bottom embedding layers, suggesting that gradient-based methods are unreliable for layer-wise analysis and identifying important tokens in the context mixing process.

\begin{table}[t]
\centering
{\small
\resizebox{0.75\linewidth}{!}{
\begin{tabular}{l c c} 
\toprule
Method & $\rho^{\text{PT}}$ & $\rho^{\text{FT}}$\\
\midrule 
Rand & -0.07~ & -0.02~ \\
\midrule
Attn & 0.10 & 0.08 \\
Attn-norm & 0.52 & 0.56\\
Attn-norm + RES & -0.35~ & -0.24~~\\
Attn-norm + RES + LN & 0.12 & 0.17\\
GlobEnc$\neg$ & -0.01~ & -0.12~~\\
ALTI$\neg$ & -0.01~ & -0.12~~\\
\midrule
GradXinput & -0.96~ & -0.99~~\\
IG & -0.86~ & -0.77~~\\
DL & -0.97~ & -1.00~~\\
\midrule
{\methodName} & \textbf{0.65} & \textbf{0.64}\\
\bottomrule
\end{tabular}
}
}
\caption{Spearman's $\rho$ correlation between layer-wise probing performance (Comp.) and layer-wise cue alignments based on Dot Product. \text{PT} and \text{FT} refer to pre-trained and fine-tuned conditions, respectively.}
\label{tab:corr_probing_rationale}
\end{table}

\section{Evaluation 3: Faithfulness Analysis}
\label{sec:faithfulness}
Our experimental results in Sections~\ref{sec:rationale_layerwise_results} and \ref{sec:probing} show that the {\methodName} score matches our prior linguistically-informed expectations. However, it is not always clear whether a {\it plausible} context mixing score that matches human expectations is also faithful to the model and  reflects its decision making process \cite{Herman2017ThePA, wiegreffe-pinter-2019-attention, jacovi-goldberg-2020-towards}.

In this section we employ the notion of input ablation \citep{Covert2021ExplainingBR} to evaluate the faithfulness of our context mixing score.
The influence of a target token on a model's decision is often estimated as the drop in the model's predicted probability of the correct class after blanking out the target token from the input. A higher drop for an ablated token indicates that the token is more influential on the model's decision \citep{deyoung-etal-2020-eraser, abnar-zuidema-2020-quantifying, atanasova-etal-2020-diagnostic, Wang2022AFI}. We use this {\it blank-out} approach as a base for analyzing and comparing the faithfulness of context mixing scores. 

To estimate the blank-out scores in BERT, we calculate the probability of its output $y$ using a softmax function normalized over only the corresponding logit values of target $t$ and foil words (cf.\ Table~\ref{Tab:blimp}), and compute blank-out scores for a given input token $i$ as $p({y_t} | \bm{e}) - p({y_t} | \bm{e} \backslash e_i)$,
where $\bm{e_i}$ refers to the input embedding of input token $i$.
We compare these blank-out scores with context mixing scores, aggregated across all layers of the model. For gradient-based scores, calculating them with respect to the tokens in the input embedding layer ($\ell=0$) provides us with aggregated scores since the backpropagation of gradients passes through all layers to the beginning of the model. For other scores, we use the \emph{rollout} \citep{abnar-zuidema-2020-quantifying} aggregation method.

Table \ref{tab:corr_blankout} shows Spearman's rank correlation between the blank-out scores and different aggregated context mixing scores. The highest correlation for our method indicates that {\methodName} is more faithful in explaining the model behaviour compared to other analysis methods.

\begin{table}[t]
\centering
{\small
\resizebox{0.75\linewidth}{!}{
\begin{tabular}{l c c} 
\toprule
Method & $\rho^{\text{PT}}$ & $\rho^{\text{FT}}$\\
\midrule 
Rand & -0.01 & ~0.00 \\
\midrule
Attn & -0.10 & -0.07 \\
Attn-norm & ~0.19 & ~0.14\\
Attn-norm + RES & ~0.03 & -0.05\\
Attn-norm + RES + LN & -0.08 & -0.17\\
GlobEnc & -0.01 & -0.09\\
ALTI & ~0.17 & ~0.19\\
\midrule
GradXinput & ~0.11 & ~0.16\\
IG & ~0.07 & ~0.21\\
DL & ~0.20 & ~0.29\\
\midrule
{\methodName} & ~\textbf{0.26} & ~\textbf{0.31}\\
\bottomrule
\end{tabular}
}
}
\caption{Spearman's $\rho$ correlation between the blank-out scores and different aggregated context mixing and attribution scores. \text{PT} and \text{FT} refer to pre-trained and fine-tuned conditions, respectively.}
\label{tab:corr_blankout}
\end{table}

\begin{figure}[t]
\centering
    \includegraphics[width=\columnwidth]{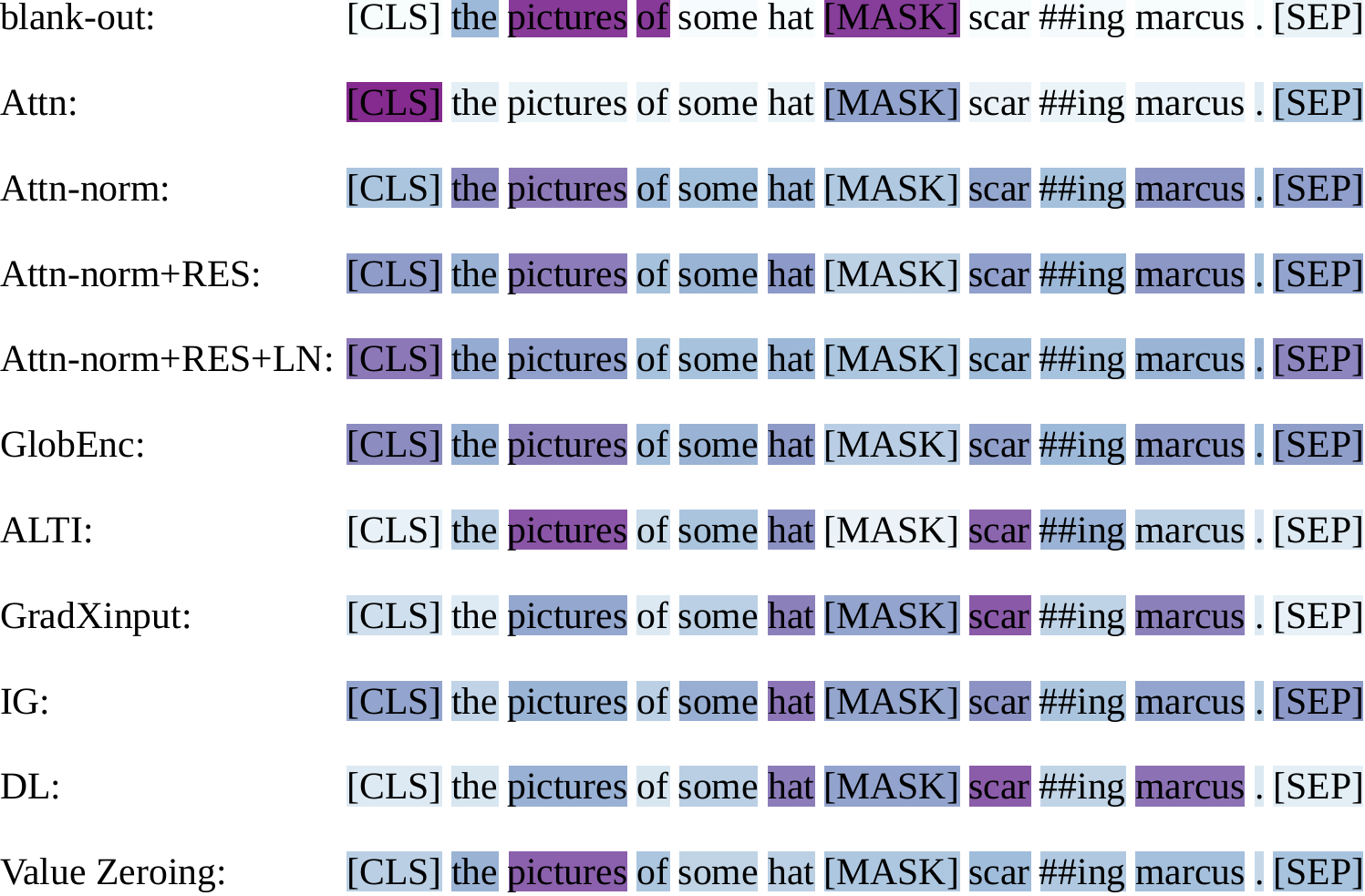}
    \caption{Most influential tokens on the target representation in a fine-tuned BERT model according to different aggregated context mixing scores compared to blank-out scores.}
    \label{fig:examplewise}
\end{figure}

\paragraph{Qualitative Analysis.}
We also take a closer look at the aggregated scores for a qualitative comparison.
In Figure~\ref{fig:examplewise}, we illustrate different scores obtained from a fine-tuned BERT model for a correctly classified example, where the model is asked to fill the masked token with one of the verbs \emph{were} or \emph{is} as target and foil classes, respectively. 
According to {\methodName} scores, the model mainly relies on the main subject (\emph{pictures}) as a cue word to form a contextualized representation of the [MASK] token, while the word \emph{pictures} is also important for the model's final decision based on the blank-out scores. In this example, the blank-out score for the cue word is $0.99$, meaning the model fully loses its confidence in the target class when the cue word is replaced with an \textsc{[UNK]} token. Surprisingly, gradient-based methods tend to highlight the word \emph{hat} which is an agreement attractor, and attention-based scores tend to focus on the \textsc{[CLS]} token which has been idle during fine-tuning process.

Overall, our faithfulness evaluation and qualitative analysis suggest that {\methodName} can explain model decisions at a global level when it is aggregated across layers. The context mixing maps per layer are provided in Appendix~\ref{appendix:layerwise_maps}, where some more meaningful patterns can be found in Value Zeroing scores (in both layer-wise and aggregated setups) in contrast to other context mixing scores.

\section{Discussion}
Although some desiderata such as plausibility \citep{lei-etal-2016-rationalizing, strout-etal-2019-human} and faithfulness \citep{10.1145/3306618.3314229, jacovi-goldberg-2020-towards} are taken into account when developing explanation and analysis methods, evaluating them is still a challenge due to lack of a standard ground truth. Evaluating context mixing scores, where token-to-token interactions in a context are also considered, is even more challenging.
Several studies have used gradient-based scores as an anchor of faithfulness, and measure how strongly context mixing scores correlate with them \citep{jain-wallace-2019-attention, abnar-zuidema-2020-quantifying, modarressi-etal-2022-globenc}. 
However, the reliability of gradient-based scores can be questioned, especially when different variations of them show considerable disagreement 
\citep{Neely2022ASO, Pruthi2022EvaluatingEH, Krishna2022TheDP}.
Thus, we suggest using controlled tasks for which we have strong prior expectations for evaluating these methods. 
In our study, we use a set of number agreement tasks to provide such priors, since the cue words are the only sources of information in the context for performing well in the task.

Another point worth discussing is the concern raised by \citet{kobayashi-etal-2021-incorporating} 
that BERT tends to preserve token representations rather than mixing them at each layer. 
We argue that their observation is due to the context-mixing ratio they defined by comparing the norm of residual effects against other token representations. In our view, this ratio is dominated by residuals and neglects the fact that a token representation carried by residual connections is indeed a contextualized representation outputted from previous layers. We keep the residuals intact within the encoder layer by zeroing only the value vectors and focusing on the context mixing performed by all tokens.

\section{Conclusion}
In this paper, we propose Value Zeroing as a novel approach for quantifying the information mixing process in Transformers to address the shortcomings of previous methods.
We performed extensive complementary experiments and showed that our method outperforms others in three different evaluation setups.
Since our approach requires no supervision, it could be an interesting option for improving model efficiency by removing token representations across layers.

\section{Limitations}
As is the case for most attempts at interpreting Deep Learning models, our evaluation of our (and others') proposed methods are not definite since we have no gold standard of what happens inside a model, although we try to remedy for that by conducting independent and complementary evaluation schemes.

Our proposed method is customized for deep neural models based on the Transformer architecture and cannot be easily generalized to other (mathematically different) modeling architectures.
Our evaluations were based on encoder-based models, and focused on the Text modality. In the future, we will extend our experiments to more modalities, such as speech and vision.

\section*{Acknowledgments}

This publication is part of the project \emph{InDeep: Interpreting Deep Learning Models for Text and Sound} (with project number NWA.1292.19.399) of National Research Agenda (NWA-ORC) programme. Funding by the Dutch Research Council (NWO) is gratefully acknowledged.

\bibliography{custom}
\bibliographystyle{acl_natbib}

\appendix
\counterwithin{figure}{section}
\counterwithin{table}{section}

\section{Appendices}

\subsection{On the choice of distance function}
\label{appendix:more_distance_functions}
\begin{figure}[th!]
\subfloat[Pre-trained BERT]{%
    \includegraphics[width=\columnwidth]{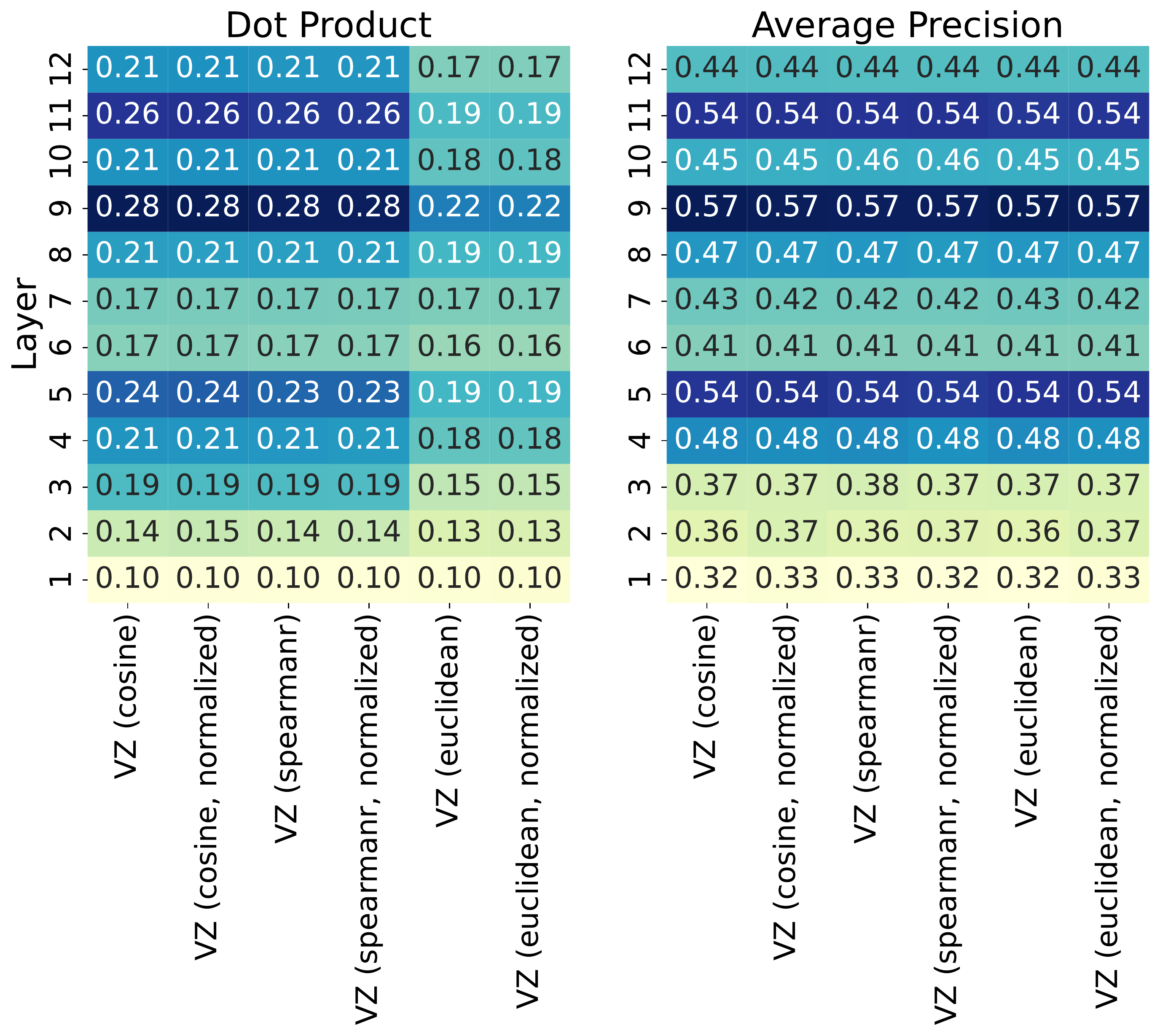}
}

\subfloat[Fine-tuned BERT]{%
    \includegraphics[width=\columnwidth]{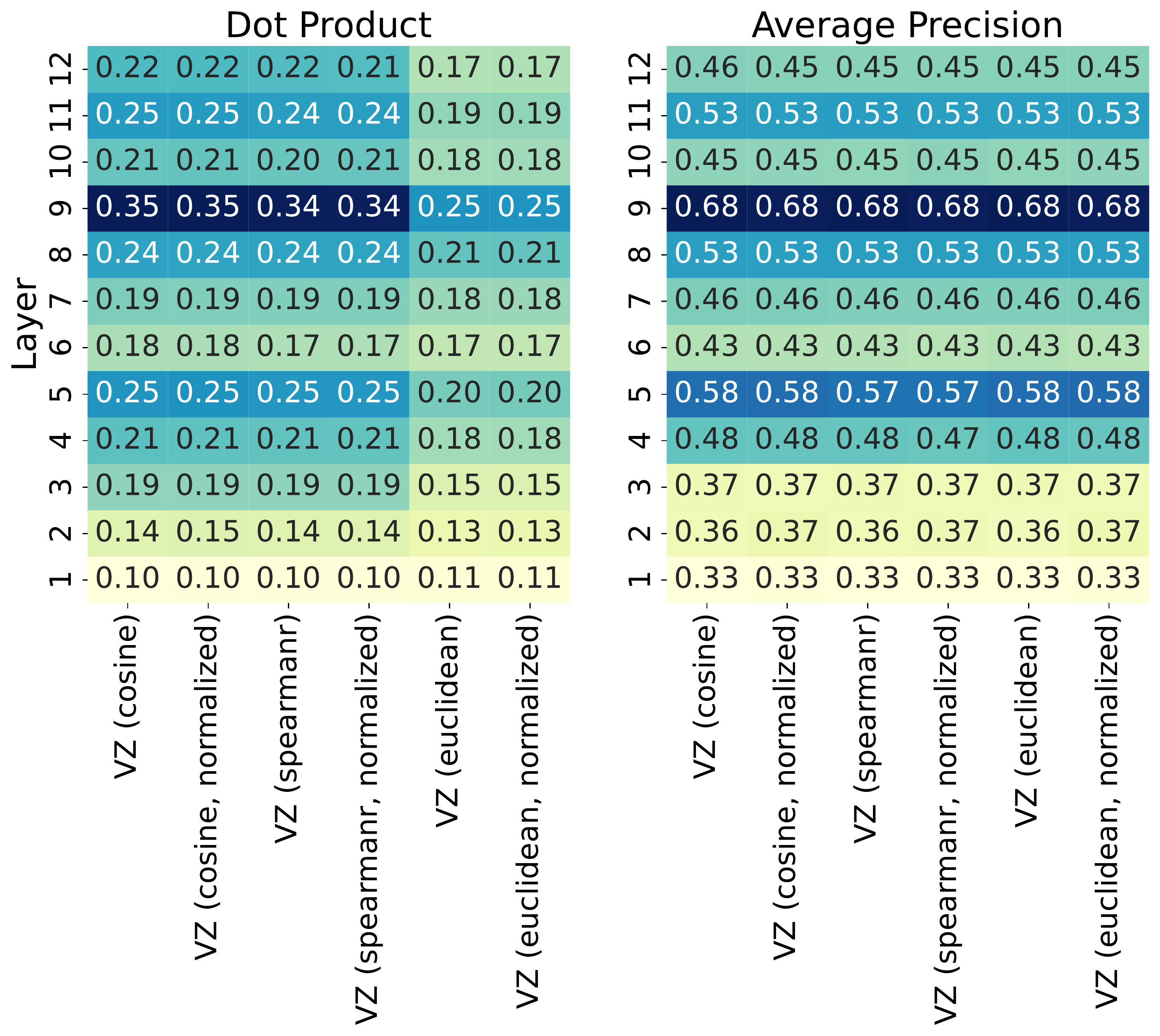}
}
\caption{Layer-wise alignment between the cue vector and different {\methodName} (VZ) scores computed based on 1) different distance metric and 2) whether representations are normalized.} 
\label{fig:compare_DFs_layerwise_rationale}
\end{figure}

The purpose of this section is to inspect the impact of selecting different distance metrics when computing Value Zeroing in Eq. \ref{eq:dependency_score}.
\citet{timkey-van-schijndel-2021-bark} questioned the informativity of standard representational distance measures such as cosine and Euclidean by observing that only a small subset of rogue dimensions contribute to the anisotropy of a contextualized representation space. They proposed using simple post-processing techniques to correct for such these rough dimensions. We followed their suggestion and normalized the representations before computing distances, but we did not observe any noticeable difference in our scores compared to using non-normalized representations (Fig.~\ref{fig:compare_DFs_layerwise_rationale}). 
We also repeated our experiment with Spearman's and Euclidean distance metrics and observed the same pattern in the results (Fig.~\ref{fig:compare_DFs_layerwise_rationale}). 

We believe that in anisotropy studies that use clustering methods, the choice of distance metrics is crucial. However, we compute each token's distance from itself (not from other tokens) and compare them relatively. This might explain why we observe the same pattern for different distance metrics.

\subsection{More metrics}
\label{appendix:more_metrics}
\begin{figure*}[h!]
\subfloat[Pre-trained BERT]{
	\includegraphics[width=\columnwidth]{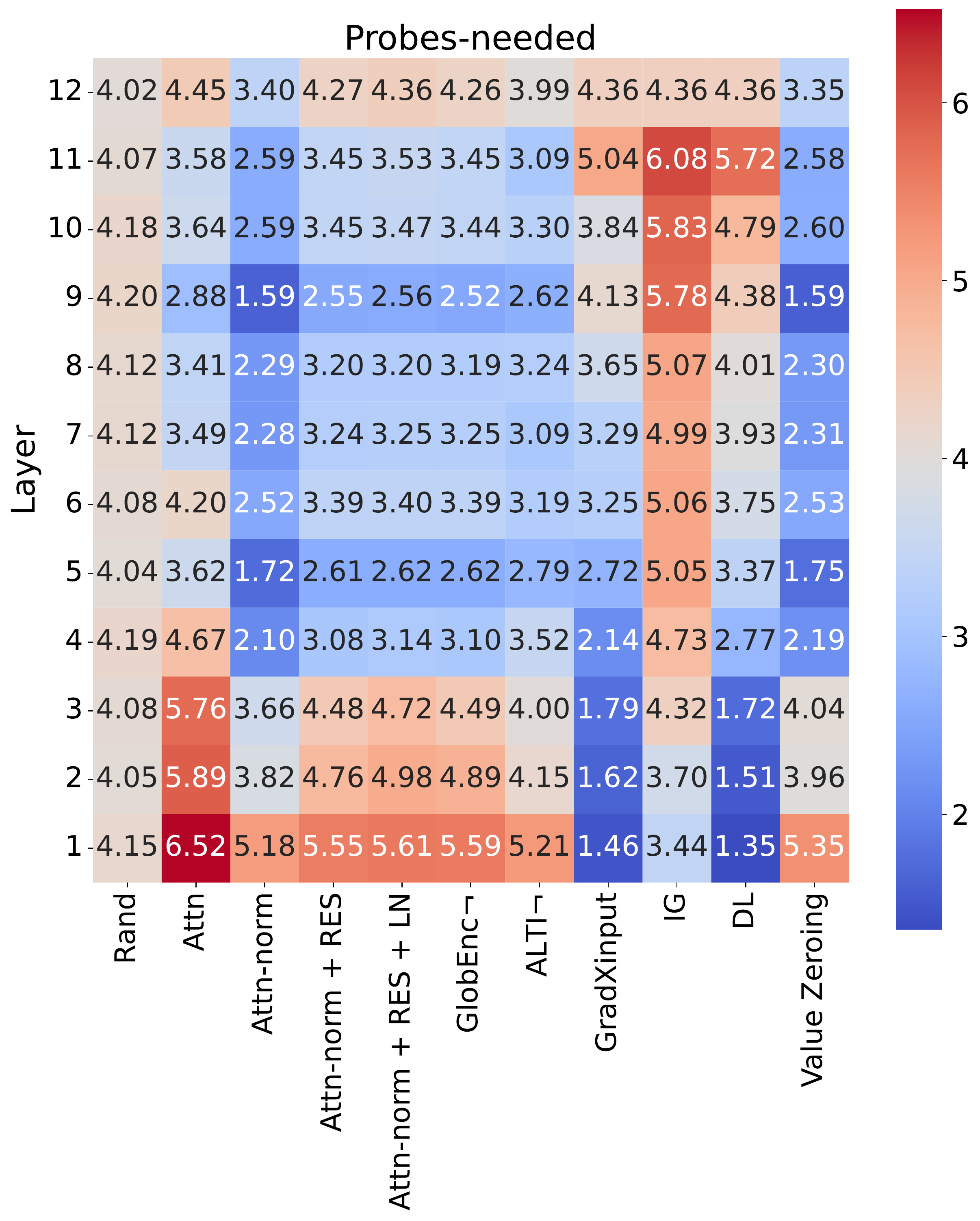} } 
\subfloat[Fine-tuned BERT]{
	\includegraphics[width=\columnwidth]{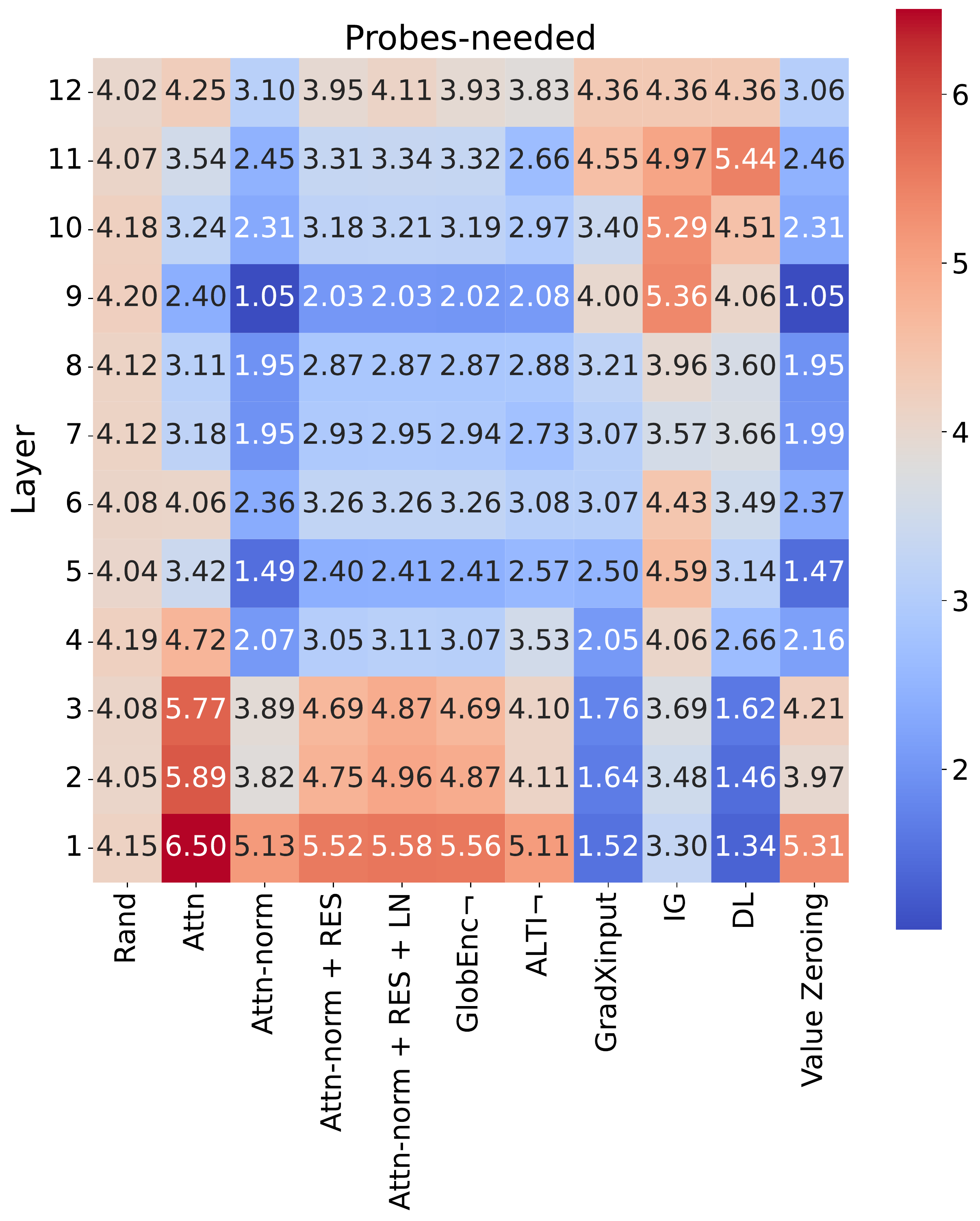} } 
\caption{The layer-wise alignment based on Probes-needed metric between the cue vector and different analysis methods averaged over Test set examples. Lower value (darker blue) is better.}
\label{fig:layerwise_rationale_pn}
\end{figure*}

Figure~\ref{fig:layerwise_rationale_pn} reports the cue alignment evaluation for BERT model based on Probes-needed \citep{Zhong2019FinegrainedSA} metric.

\subsection{More PLMs}
\label{appendix:more_plms}
\begin{figure*}[h!]
\subfloat[Pre-trained RoBERTa]{
    \includegraphics[width=\linewidth]{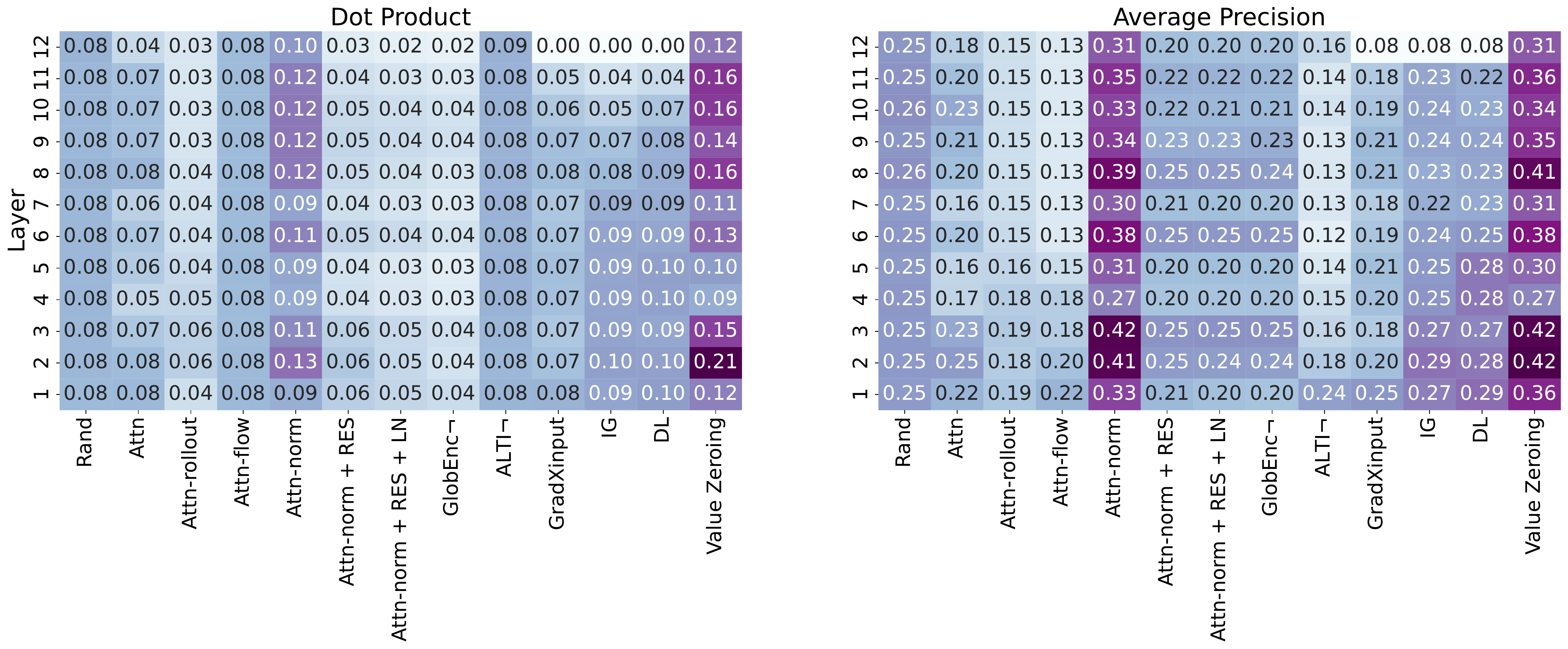}
 }
 
 \subfloat[Fine-tuned RoBERTa]{
    \includegraphics[width=\linewidth]{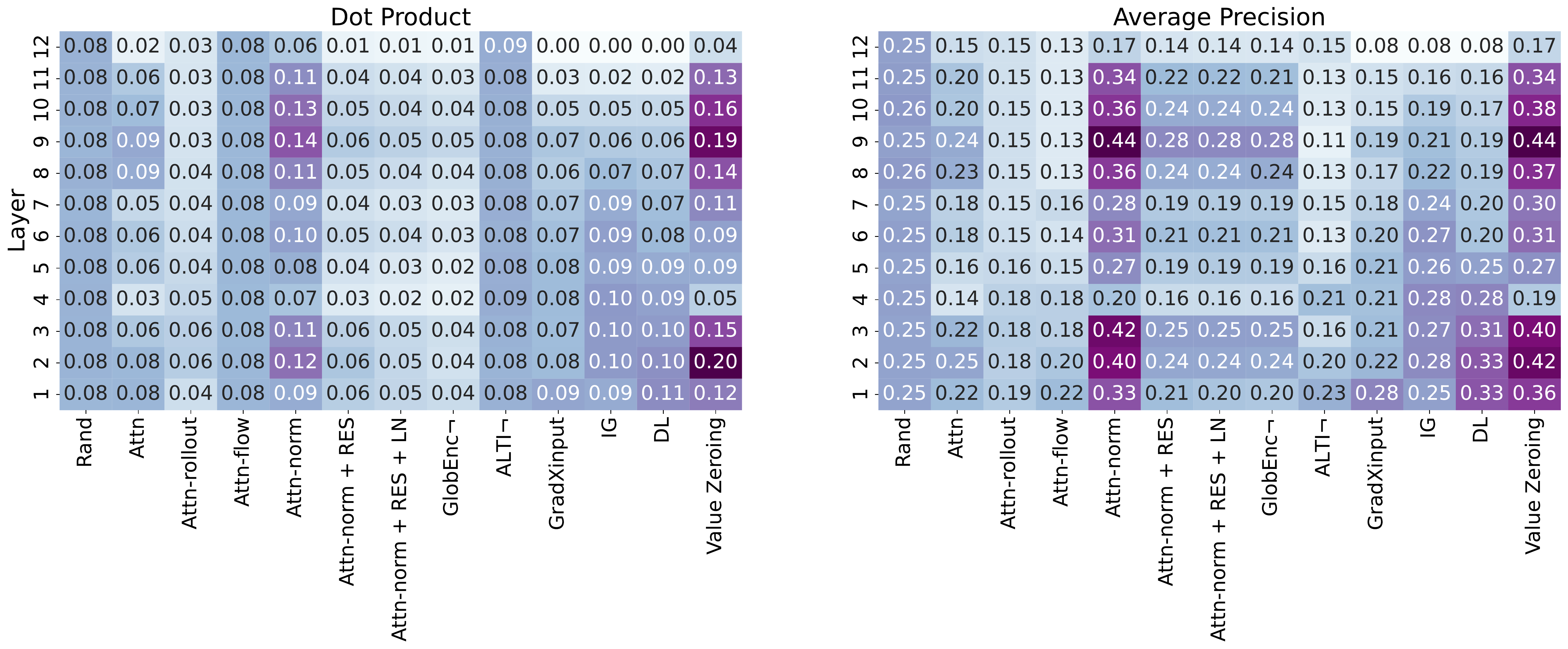}
}
\caption{Layer-wise alignment between the cue vector and different analysis methods averaged over Test set examples for RoBERTa. Higher value (darker color) is better.}
\label{fig:layerwise_rationales_roberta}
\end{figure*}

\begin{figure*}[h!]
\subfloat[Pre-trained ELECTRA]{
    \includegraphics[width=\linewidth]{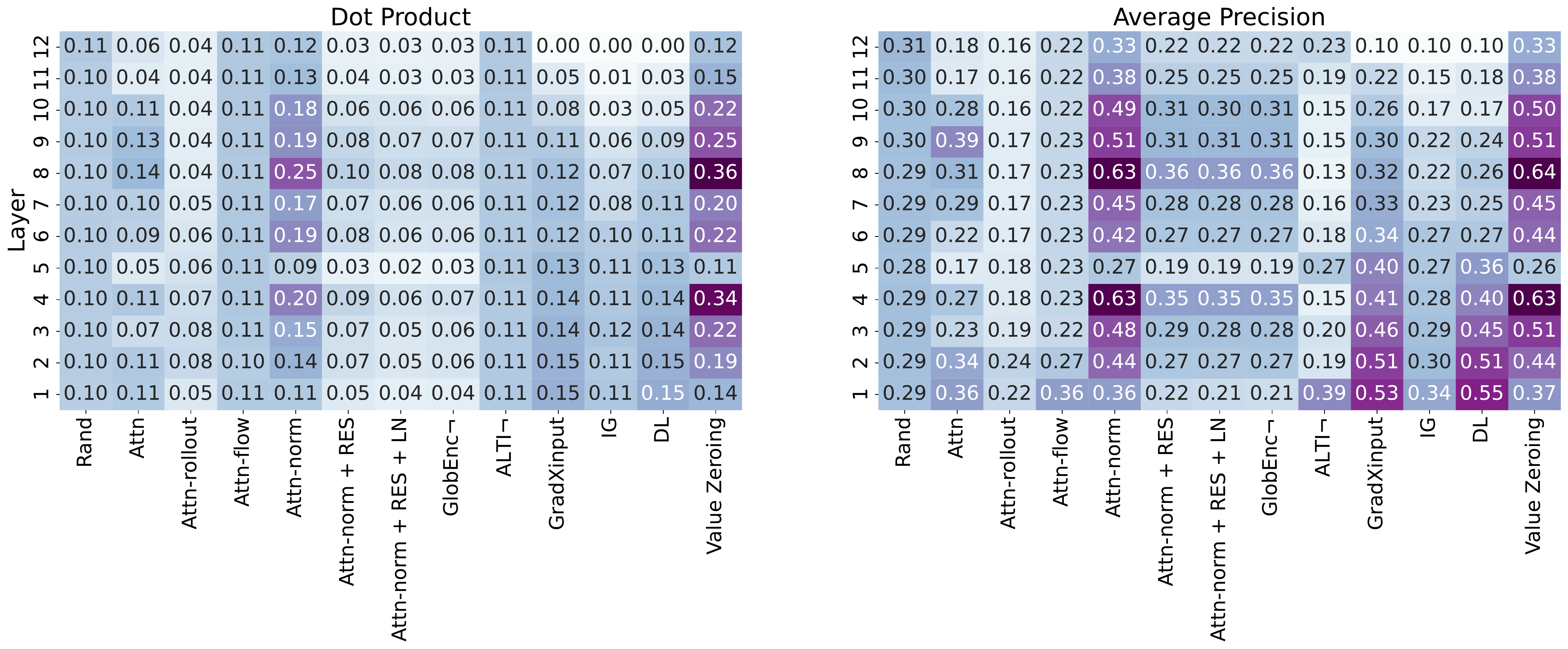}
 }
 
 \subfloat[Fine-tuned ELECTRA]{
    \includegraphics[width=\linewidth]{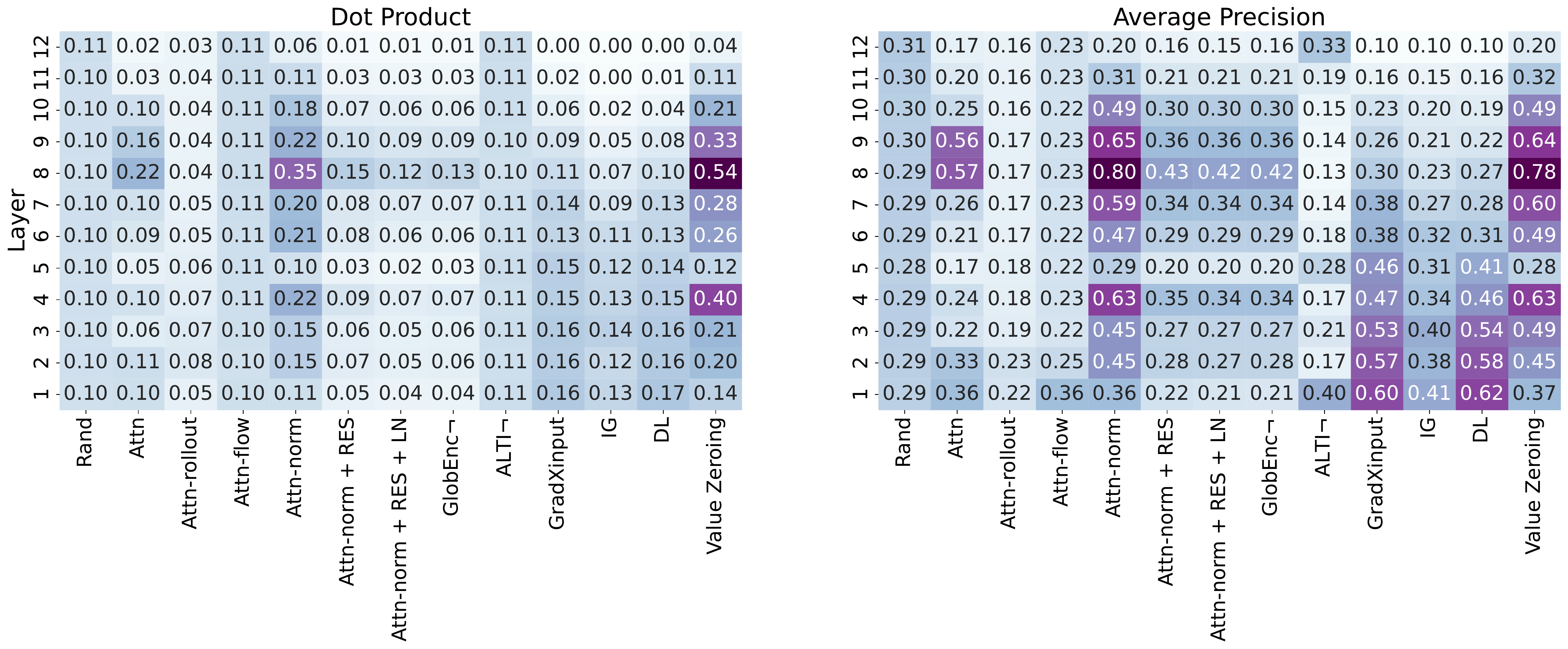}
}
\caption{Layer-wise alignment between the cue vector and different analysis methods averaged over Test set examples for ELECTRA. Higher value (darker color) is better.}
\label{fig:layerwise_rationales_electra}
\end{figure*}

We replicated our experiment for the cue alignment for two more PLMs; RoBERTa \citep{Liu2019RoBERTaAR} and ELECTRA (generator, \citealp{Clark2020ELECTRA:}). As we can see in Figures \ref{fig:layerwise_rationales_roberta} and \ref{fig:layerwise_rationales_electra}, our method consistently outperforms other methods on all models in both pre-trained and fine-tuned setups.
Due to the fact that our scores are based on zeroing value vectors, our method can be easily applied to any Transformer-based models even with different modalities.

\subsection{Qualitative Analysis: Layer-wise Context Mixing Maps}
\label{appendix:layerwise_maps}
This section illustrates different context mixing maps obtained from a fine-tuned BERT model for the correctly classified example of \emph{``The pictures of some hat [MASK] scaring Marcus.''}

\begin{figure*}[ht]
\centering
    \textbf{Attn}
    \includegraphics[width=.95\linewidth]{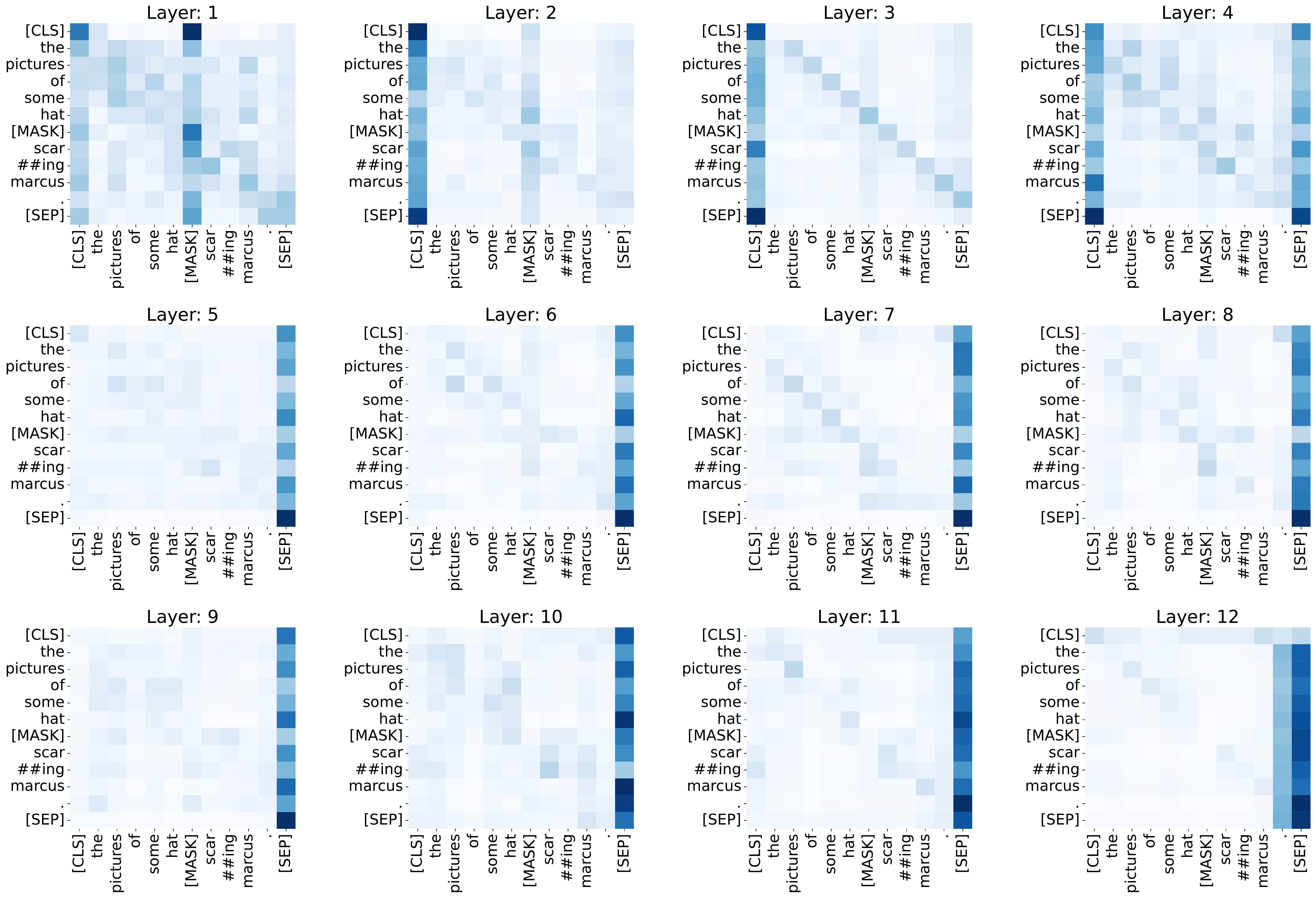}
    \label{fig:maps_ft_Attn_8}
    \caption{Raw attention scores (Attn) averaged over all attention heads at each different layer.}
\end{figure*}
\begin{figure*}[ht]
\centering
    \textbf{Attn w/ rollout}
    \includegraphics[width=.95\linewidth]{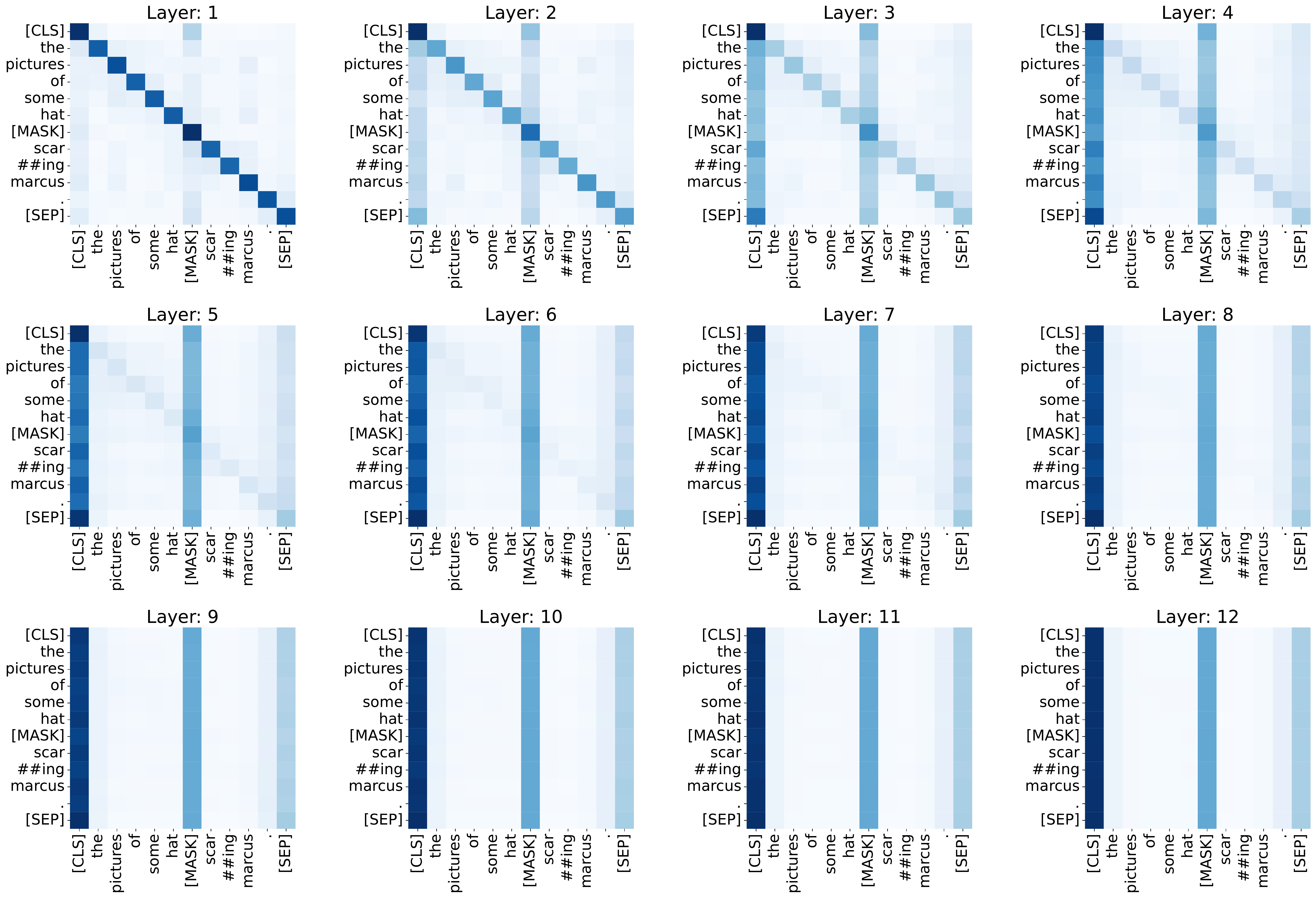}
    \caption{Raw attention scores (Attn) aggregated by rollout (\citet{abnar-zuidema-2020-quantifying}'s method) across layers.}
\end{figure*}

\begin{figure*}[ht]
\centering
    \textbf{Attn-norm}
    \includegraphics[width=.95\linewidth]{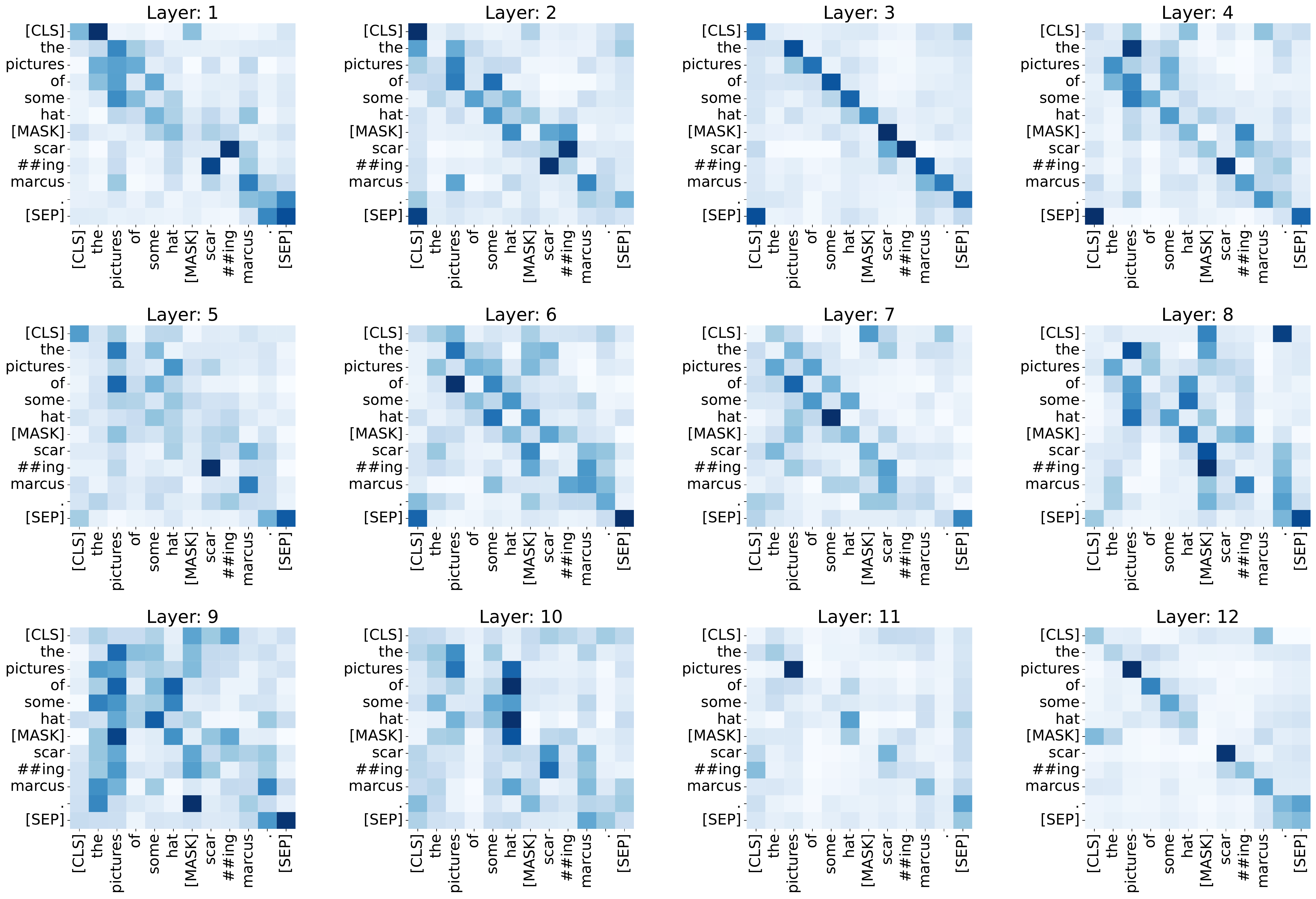}
    \caption{\citet{kobayashi-etal-2020-attention}'s scores (Attn-norm) across layers.}
\end{figure*}
\begin{figure*}[ht]
\centering
    \textbf{Attn-norm w/ rollout}
    \includegraphics[width=.95\linewidth]{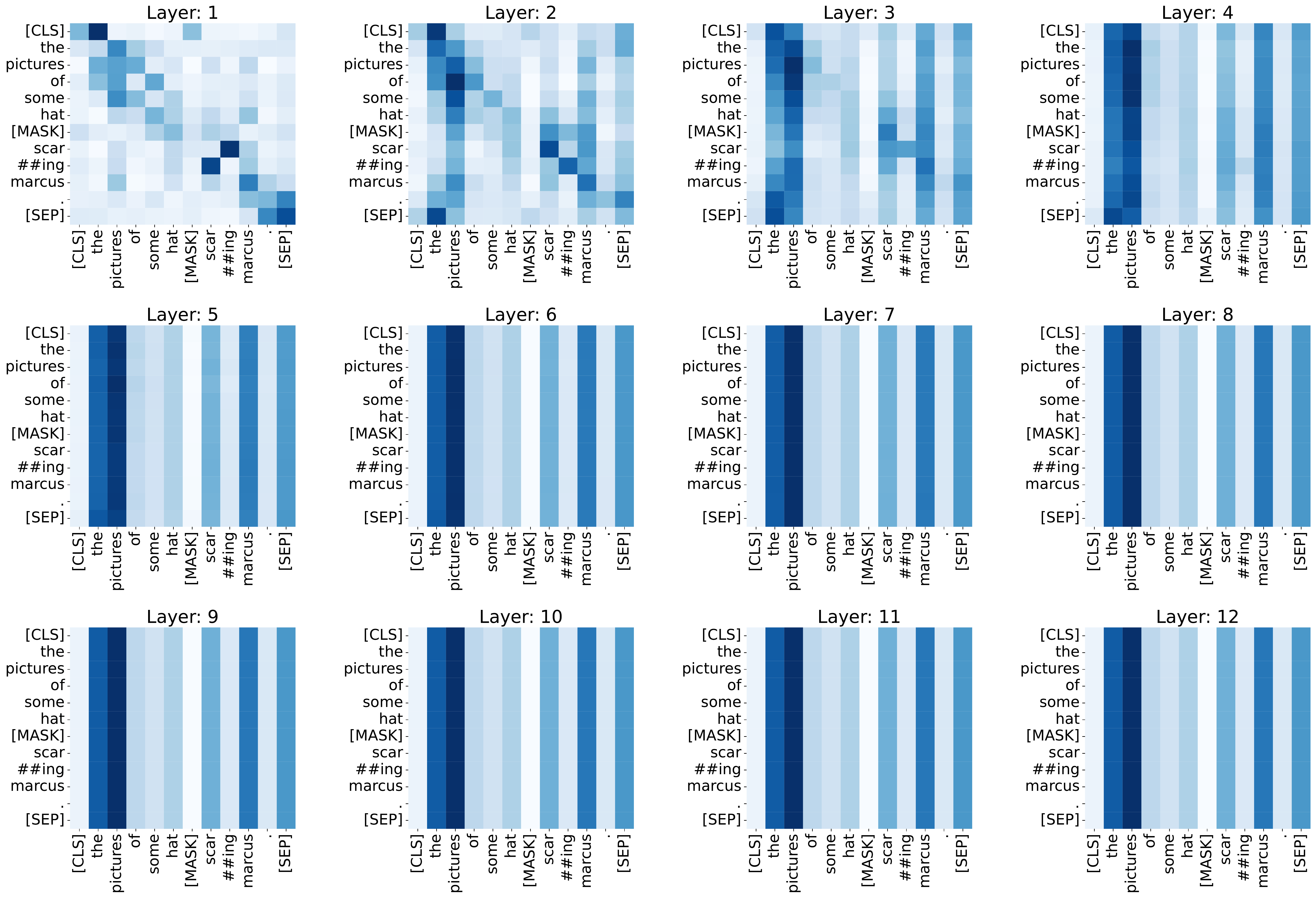}
    \caption{\citet{kobayashi-etal-2020-attention}'s scores (Attn-norm) aggregated by rollout method across layers.}
\end{figure*}

\begin{figure*}[ht]
\centering
    \textbf{Attn-norm + RES}
    \includegraphics[width=.95\linewidth]{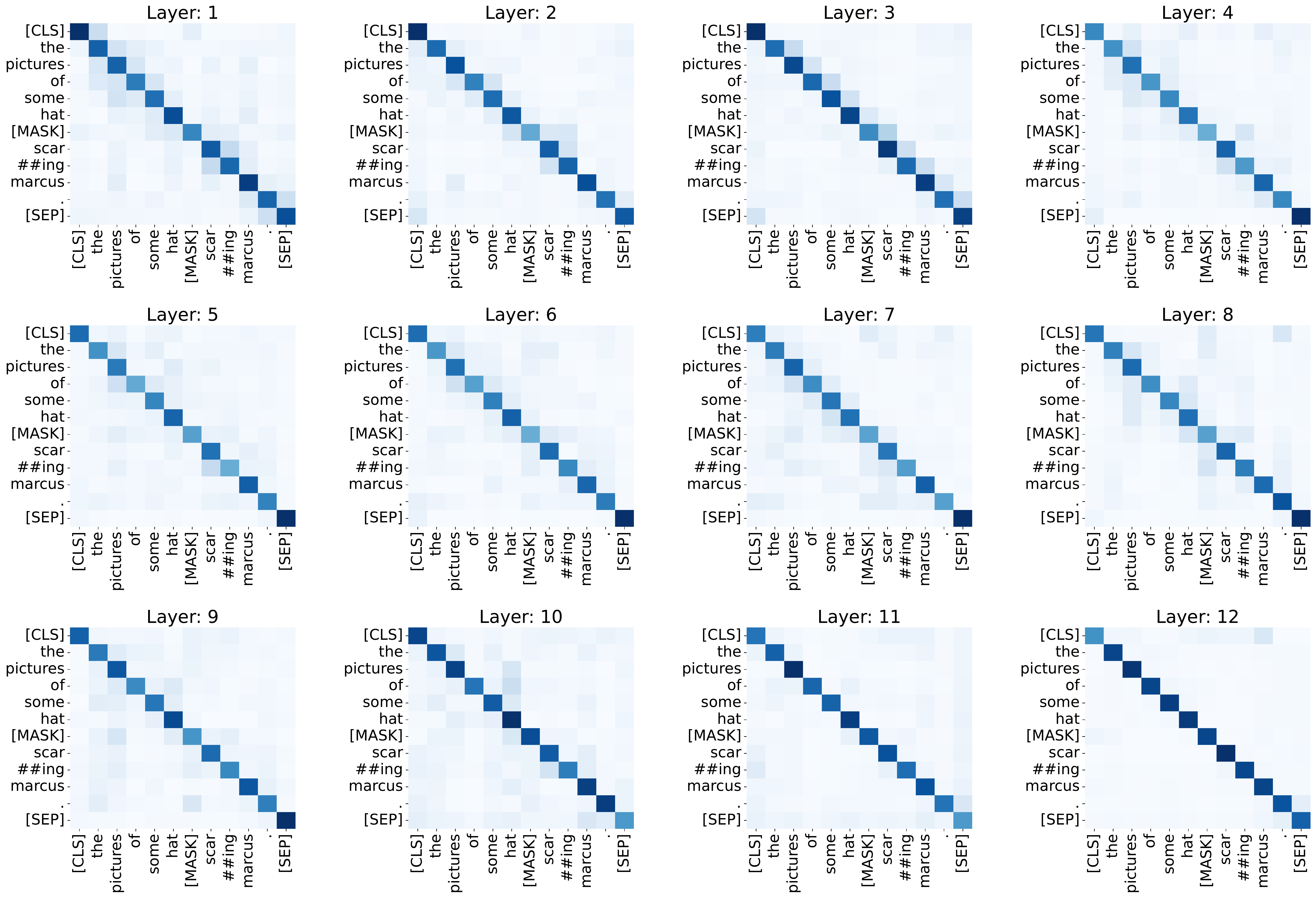}
    \caption{\citet{kobayashi-etal-2021-incorporating}'s scores (Attn-norm + RES) across layers.}
\end{figure*}
\begin{figure*}[ht]
\centering
    \textbf{Attn-norm + RES w/ rollout}
    \includegraphics[width=.95\linewidth]{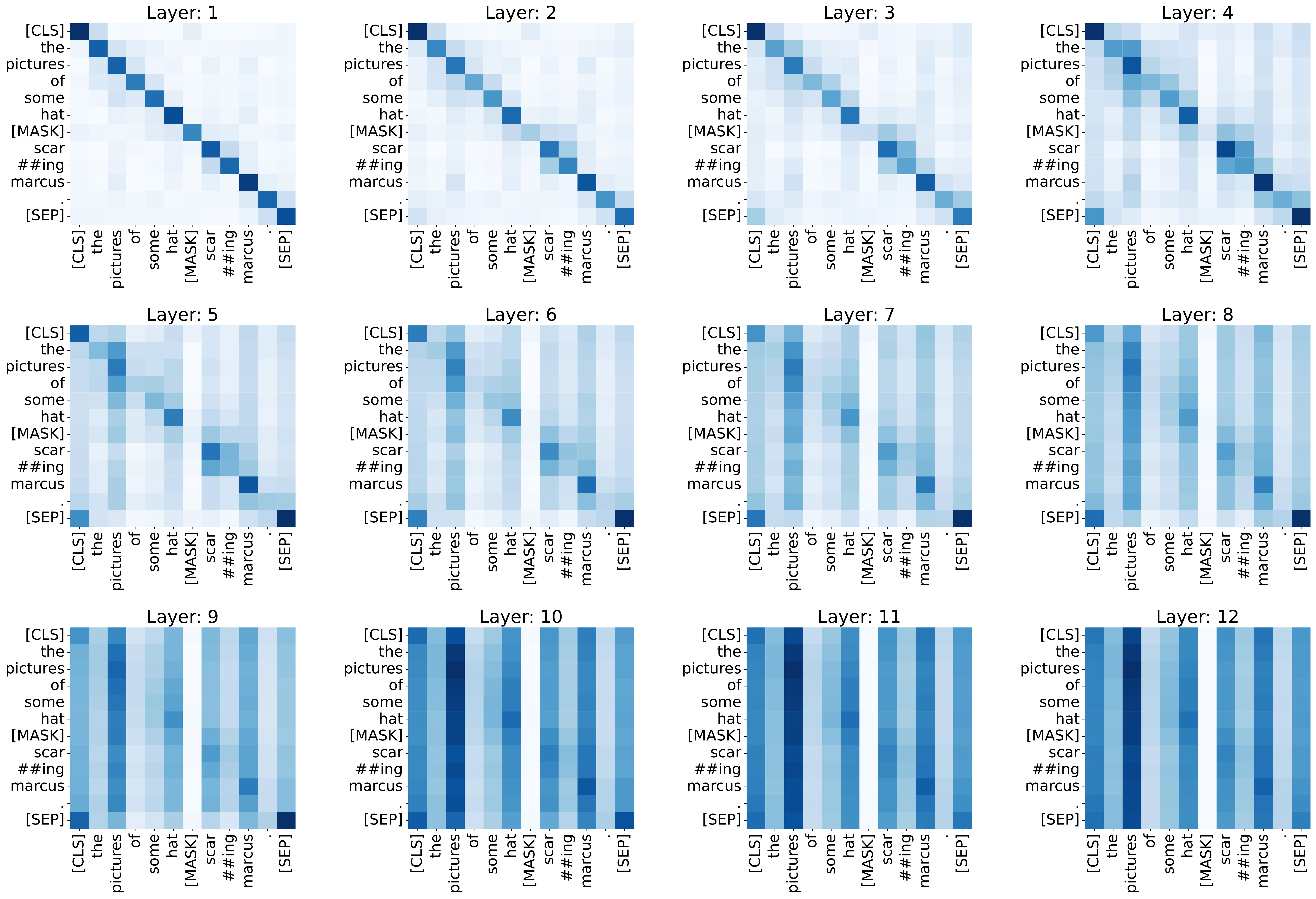}
    \caption{\citet{kobayashi-etal-2021-incorporating}'s scores (Attn-norm + RES) aggregated by rollout method across layers.}
\end{figure*}

\begin{figure*}[ht]
\centering
    \textbf{Attn-norm + RES + LN}
    \includegraphics[width=.95\linewidth]{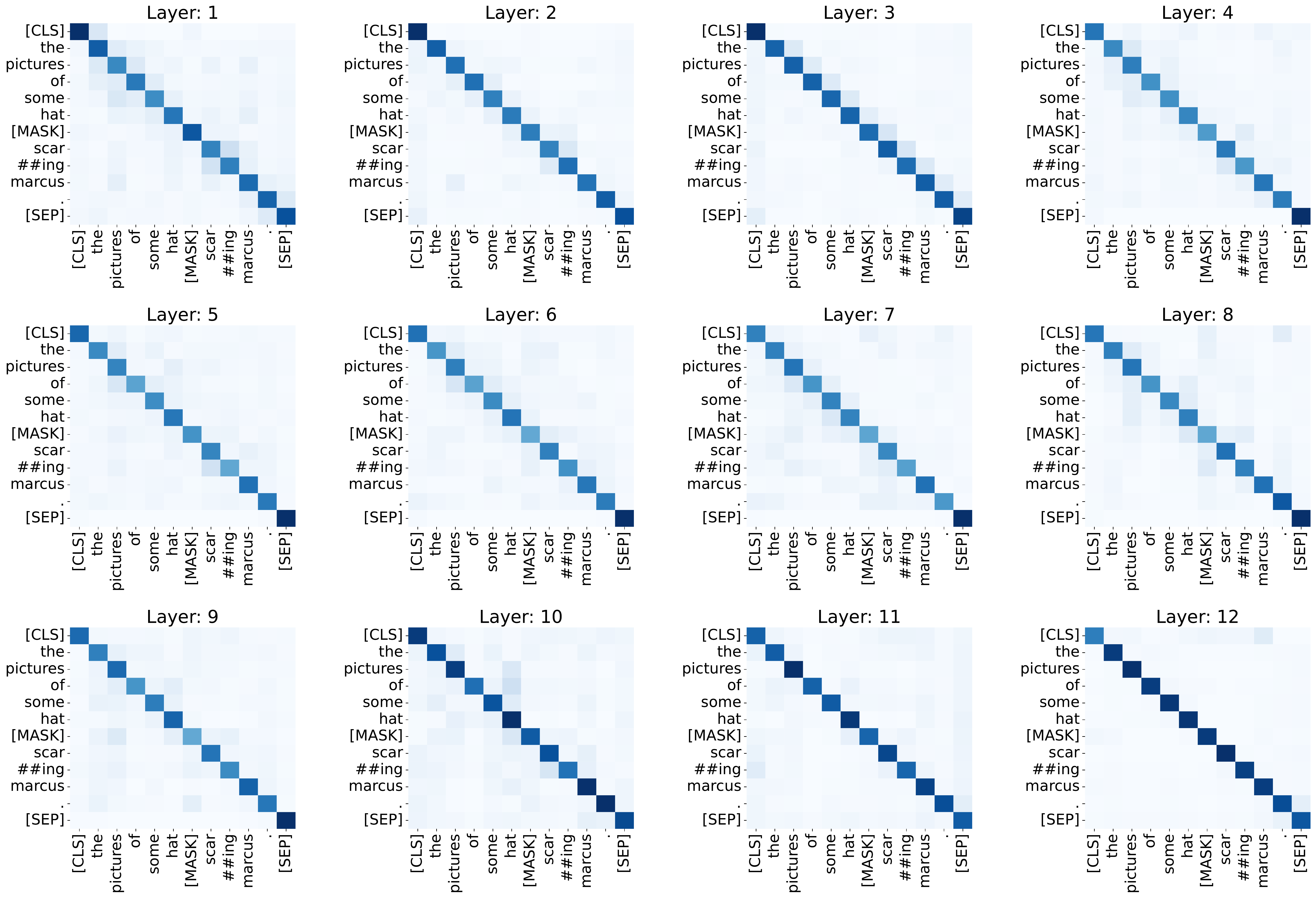}
    \caption{\citet{kobayashi-etal-2021-incorporating}'s scores (Attn-norm + RES + LN) across layers.}
    \label{fig:maps_ft_Attn-norm+RES+LN_8}
\end{figure*}
\begin{figure*}[ht]
\centering
    \textbf{Attn-norm + RES + LN w/ rollout}
    \includegraphics[width=.95\linewidth]{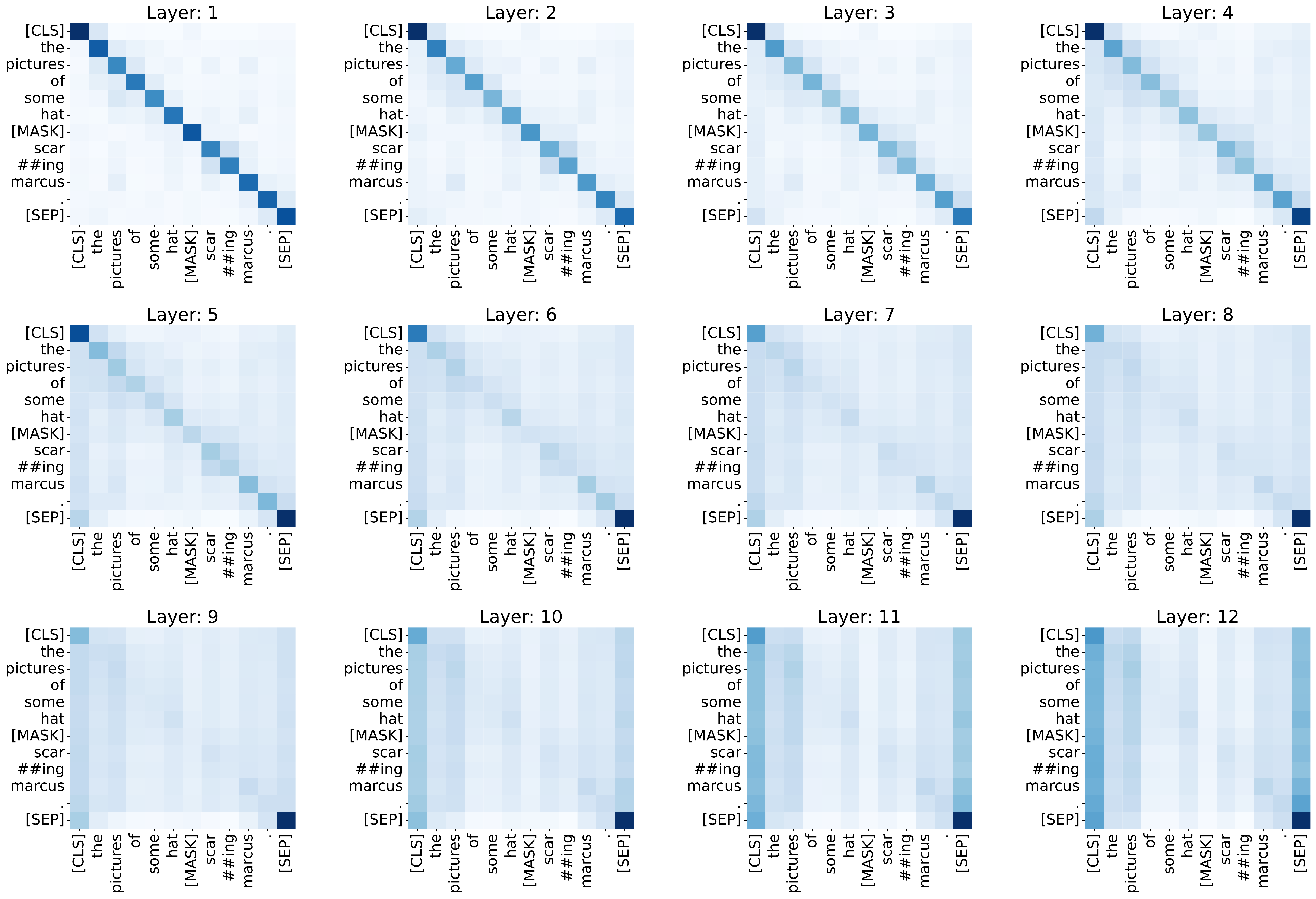}
    \caption{\citet{kobayashi-etal-2021-incorporating}'s scores (Attn-norm + RES + LN) aggregated by rollout method across layers.}
\end{figure*}

\begin{figure*}[ht]
\centering
    \textbf{GlobEnc without rollout}
    \includegraphics[width=.95\linewidth]{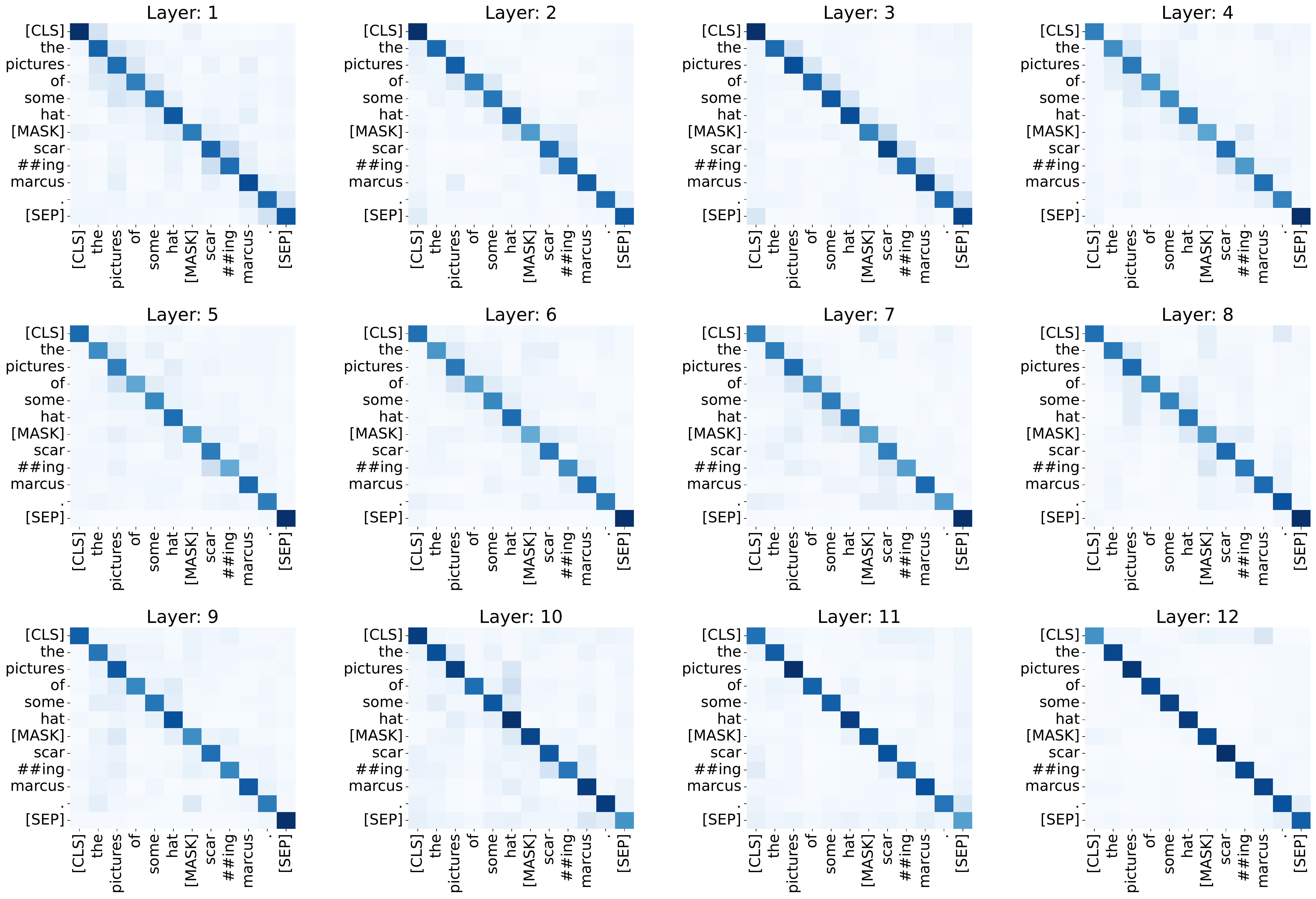}
    \caption{\citet{modarressi-etal-2022-globenc}'s scores (GlobEnc$\neg$) without aggregation (rollout) across layers.}
\end{figure*}
\begin{figure*}[ht]
\centering
    \textbf{GlobEnc (w/ rollout)}
    \includegraphics[width=.95\linewidth]{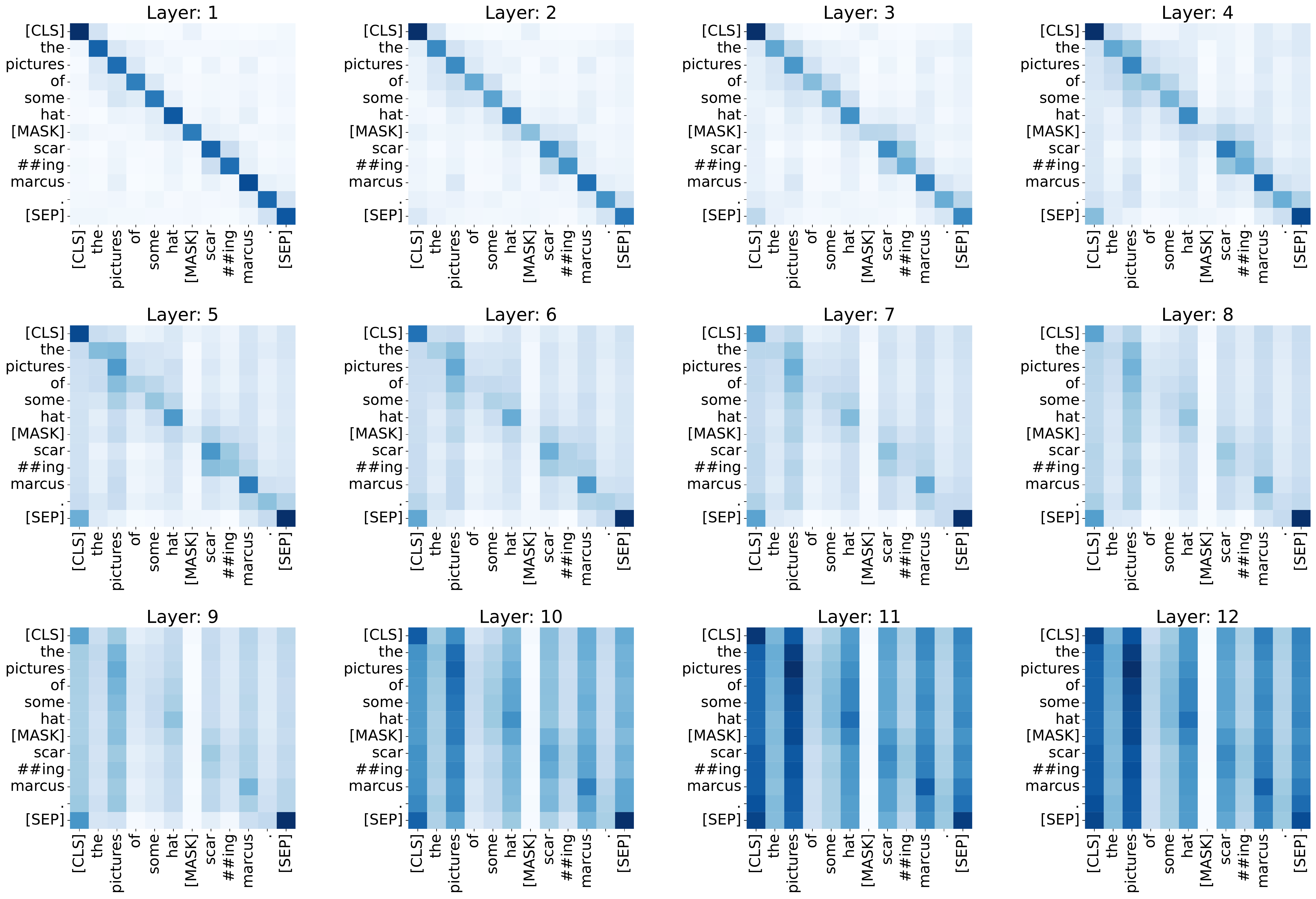}
    \caption{\citet{modarressi-etal-2022-globenc}'s scores (GlobEnc) across layers (the rollout method is inherently incorporated in GlobEnc).}
\end{figure*}

\begin{figure*}[ht]
\centering
    \textbf{ALTI without rollout}
    \includegraphics[width=.95\linewidth]{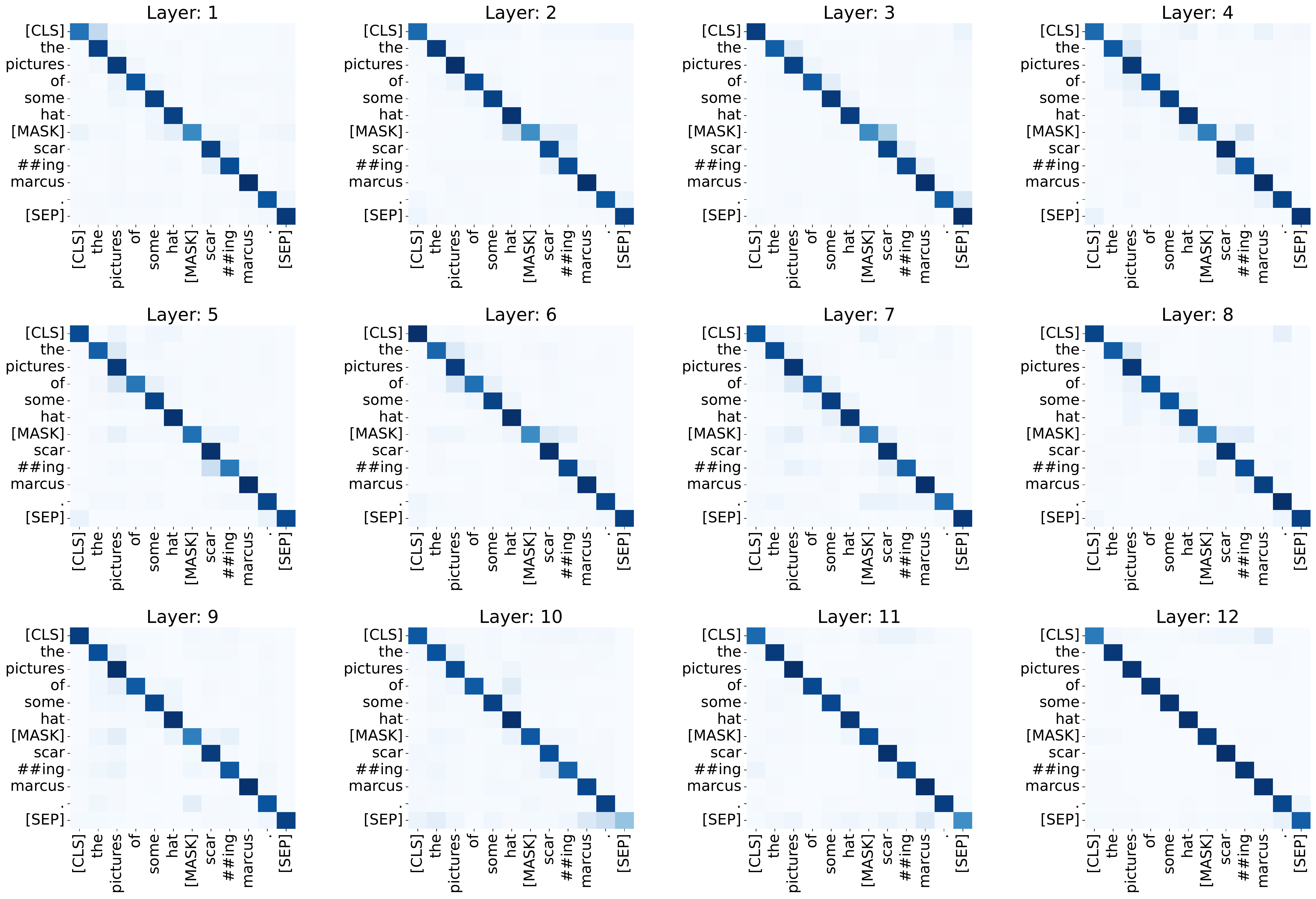}
    \caption{\citet{ferrando-etal-2022-measuring}'s scores (ALTI$\neg$) without aggregation (rollout) across layers.}
\end{figure*}
\begin{figure*}[ht]
\centering
    \textbf{ALTI (w/ rollout)}
    \includegraphics[width=.95\linewidth]{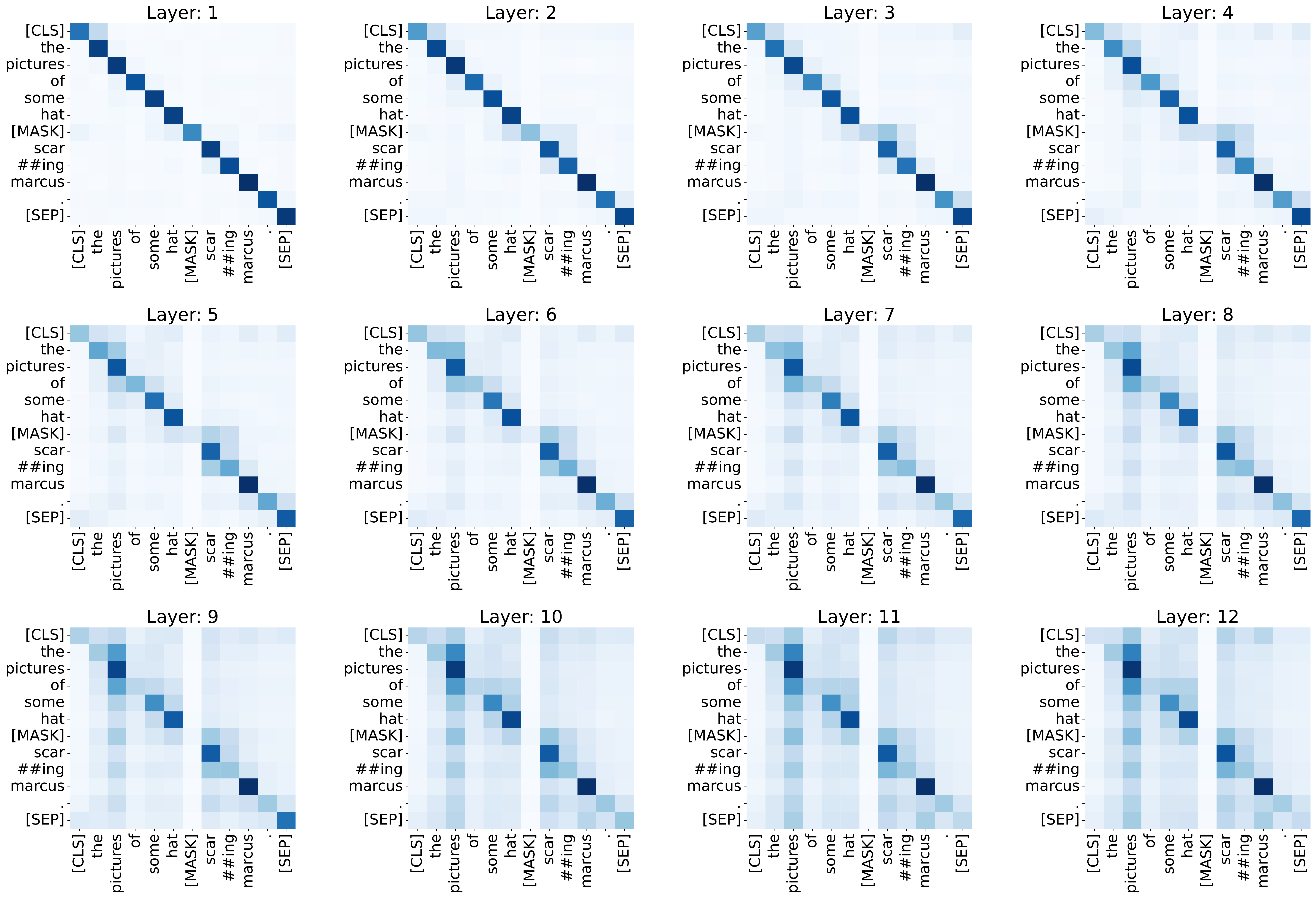}
    \caption{\citet{ferrando-etal-2022-measuring}'s scores (ALTI) across layers (the rollout method is inherently incorporated in ALTI).}
\end{figure*}

\begin{figure*}[ht]
\centering
    \textbf{Value Zeroing}
    \includegraphics[width=.95\linewidth]{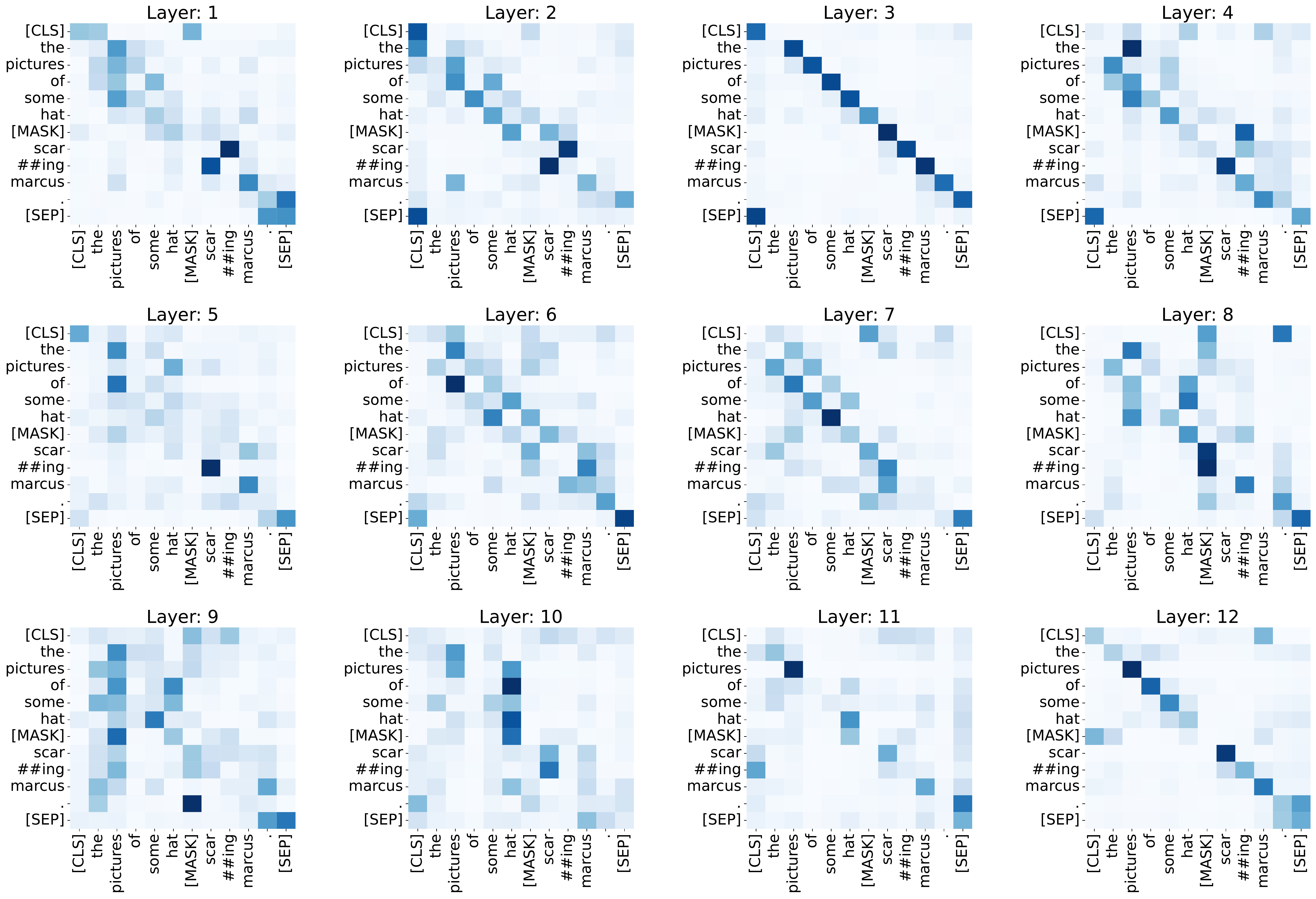}
    \caption{Our scores ({\methodName}) across layers.}
\end{figure*}
\begin{figure*}[ht]
\centering
    \textbf{Value Zeroing w/ rollout}
    \includegraphics[width=.95\linewidth]{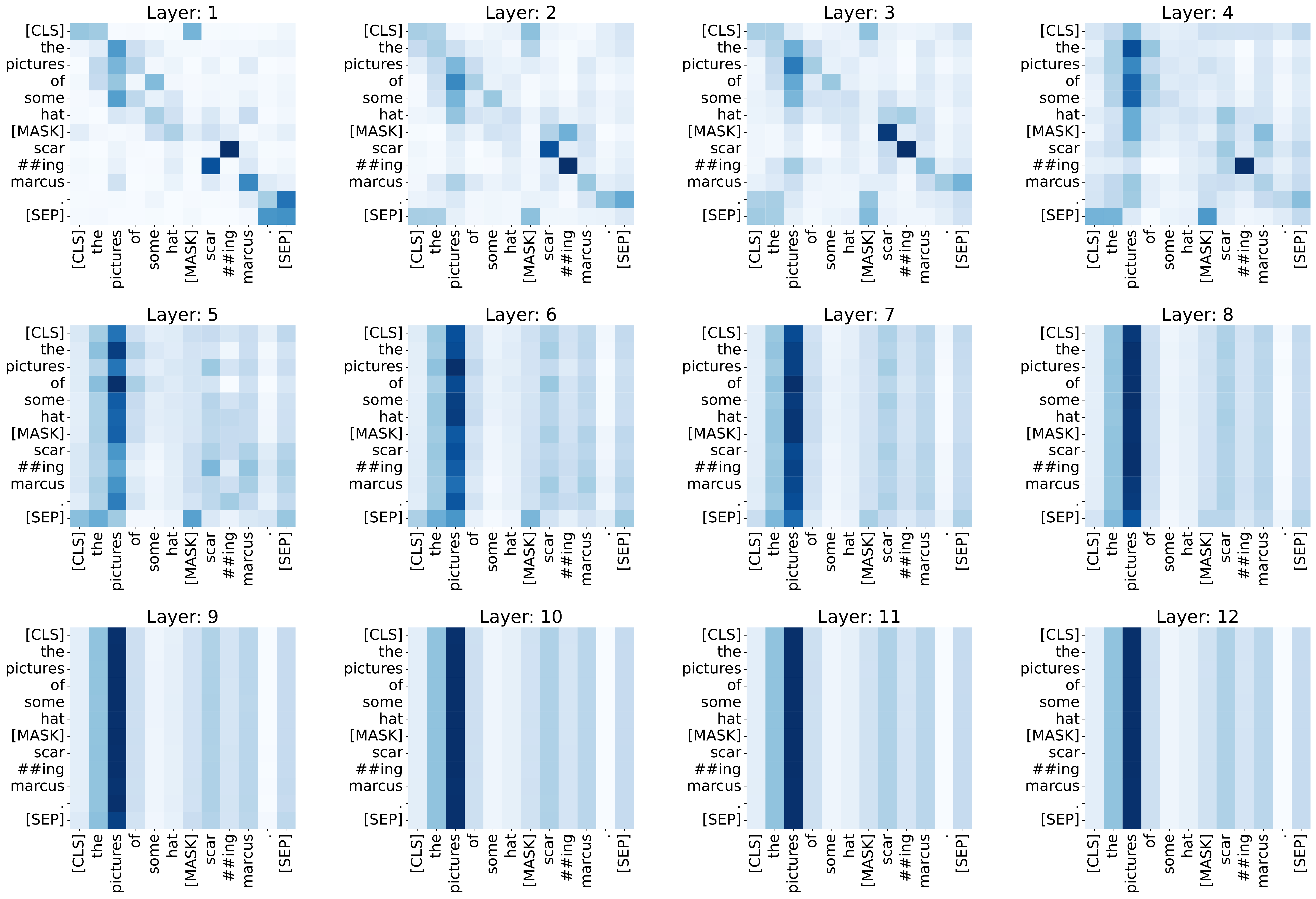}
    \caption{Our global scores ({\methodName}) aggregated by rollout method across layers.}
\end{figure*}

\end{document}